\documentclass[10pt,twocolumn,letterpaper]{article}

\usepackage{iccv}
\usepackage{times}
\usepackage{epsfig}
\usepackage{graphicx}
\usepackage{amsmath}
\usepackage{amssymb}
\usepackage{multirow}
\usepackage{subfigure}
\usepackage[pagebackref=true,breaklinks=true,letterpaper=true,colorlinks,bookmarks=false]{hyperref}

\iccvfinalcopy % *** Uncomment this line for the final submission

\def\iccvPaperID{7325} % *** Enter the ICCV Paper ID here

\ificcvfinal\fi

\begin{document}

%%%%%%%%% TITLE
\title{Distributed bundle adjustment with block-based sparse matrix compression for super large scale datasets}

\author{Maoteng Zheng\\
China University of Geosciences(Wuhan)\\
{\tt\small zhengmaoteng@cug.edu.cn}
\and
Nengcheng Chen\\
China University of Geosciences(Wuhan)\\
{\tt\small chennengcheng@cug.edu.cn}
\and
Junfeng Zhu\\
Mirauge3D Technology\\
{\tt\small jf\underline{\space}zhu@smartmappingtek.com}
\and
Xiaoru Zeng\\
Mirauge3D Technology\\
{\tt\small xr\underline{\space}zeng@smartmappingtek.com}
\and
Huanbin Qiu\\
Jiantong Surveying\\
{\tt\small  mrqq18@hotmail.com}
\and
Yuyao Jiang\\
China University of Geosciences(Wuhan)\\
{\tt\small yuyaojiang@cug.edu.cn}
\and
Xingyue Lu\\
China University of Geosciences(Wuhan)\\
{\tt\small xingyuelu@cug.edu.cn}
\and
Hao Qu\\
Mirauge3D Technology Inc.\\
{\tt\small quhao@smartmappingtek.com}
}
\maketitle
\ificcvfinal\thispagestyle{empty}\fi

%%%%%%%%% ABSTRACT
\begin{abstract}
    We propose a distributed bundle adjustment (DBA) method using the exact Levenberg-Marquardt (LM) algorithm for super large-scale datasets. Most of the existing methods partition the global map to small ones and conduct bundle adjustment in the submaps. In order to fit the parallel framework, they use approximate solutions instead of the LM algorithm. However, those methods often give sub-optimal results. Different from them, we utilize the exact LM algorithm to conduct global bundle adjustment where the formation of the reduced camera system (RCS) is actually parallelized and executed in a distributed way. To store the large RCS, we compress it with a block-based sparse matrix compression format (BSMC), which fully exploits its block feature. The BSMC format also enables the distributed storage and updating of the global RCS. The proposed method is extensively evaluated and compared with the state-of-the-art pipelines using both synthetic and real datasets. Preliminary results demonstrate the efficient memory usage and vast scalability of the proposed method compared with the baselines. For the first time, we conducted parallel bundle adjustment using LM algorithm on a real datasets with 1.18 million images and a synthetic dataset with 10 million images (about 500 times that of the state-of-the-art LM-based BA) on a distributed computing system.
\end{abstract}

\begin{figure*}[htbp]
\begin{center}
\begin{minipage}{0.99\linewidth}
\includegraphics[width=3.3cm, height=2.5cm]{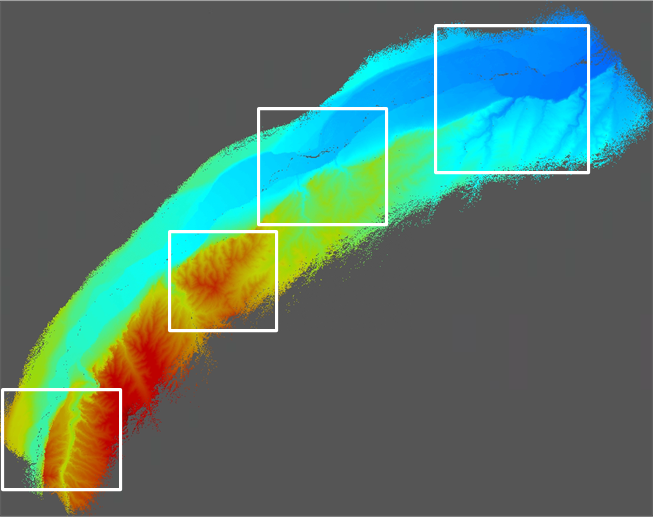}
\includegraphics[width=3.3cm, height=2.5cm]{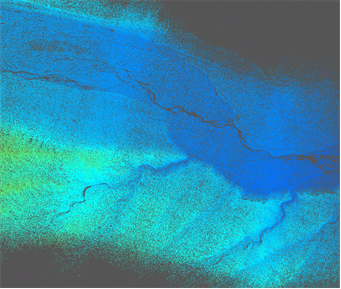}
\includegraphics[width=3.3cm, height=2.5cm]{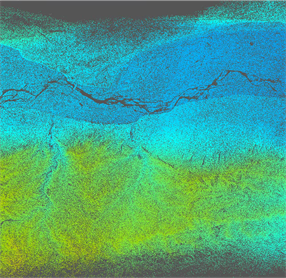}
\includegraphics[width=3.3cm, height=2.5cm]{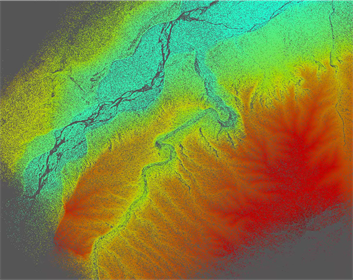}
\includegraphics[width=3.3cm, height=2.5cm]{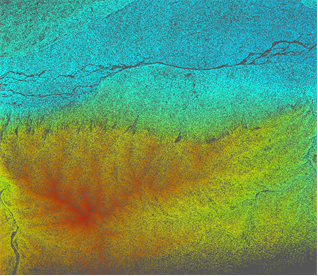}
\end{minipage}
\begin{minipage}{0.99\linewidth}
\includegraphics[width=3.3cm, height=2.5cm]{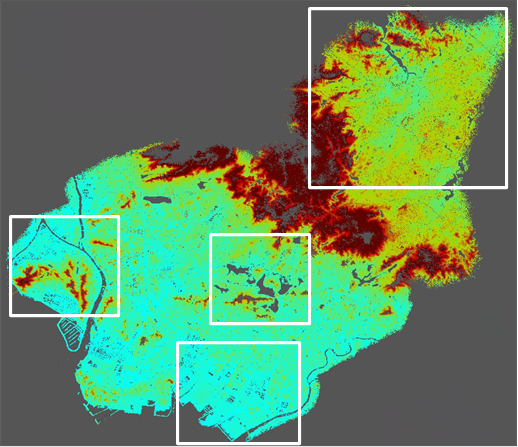}
\includegraphics[width=3.3cm, height=2.5cm]{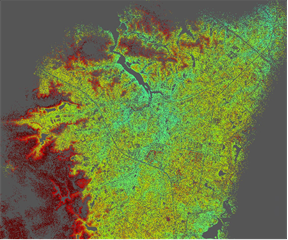}
\includegraphics[width=3.3cm, height=2.5cm]{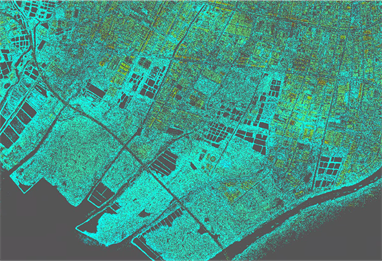}
\includegraphics[width=3.3cm, height=2.5cm]{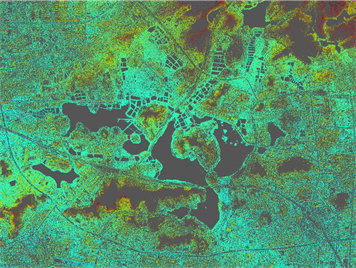}
\includegraphics[width=3.3cm, height=2.5cm]{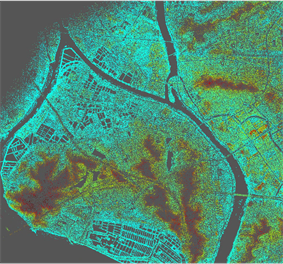}
\end{minipage}
\end{center}
   \caption{The sparse point cloud of a real dataset with 1.18 million images processed using the proposed method. The color from blue to red represents the height from low to high. The white rectangles on the left image are enlarged and shown on the right side}
\label{fig1}
\end{figure*}

%%%%%%%%% BODY TEXT
\section{Introduction}
\label{section1}
Bundle adjustment (BA) is a significant step for 3D reconstruction in both computer vision and photogrammetry communities. Its main objective is to recover the camera poses as well as the 3D points by minimizing the square sum of the reprojection errors. This nonlinear optimization problem is an ancient subject, having been studied for decades. Classic solutions such as Steepest descent, Gauss-Newton, and Levenberg-Marquardt (LM) \cite{levenberg1944method, marquardt1963algorithm} have been proposed and widely applied, and plenty of libraries have been open sourced, such as Ceres \cite{triggs2000bundle}, g2o \cite{grisetti2011g2o}, SBA \cite{lourakis2009sba}, PBA \cite{wu2011multicore}, MegBA \cite{ren2022megba}, and so on. Among them, the LM algorithm is most widely supported. In a bundle adjustment, the nonlinear system is firstly linearized at an initial guess, which is a given prior, and then solved via direct inversion, Cholesky decomposition, or preconditioned conjugate gradient (PCG) \cite{byrod2009bundle,byrod2010conjugate}.  

The Hessian matrix in a linear system is usually reduced by the Shur complement trick, forming a Reduced Camera System (RCS) that is much smaller than the Hessian. As the number of images increases, even the memory requirement for an RCS is a challenging issue. The formation, storage and inversion of the RCS gradually become the bottleneck of the bundle adjustment for large-scale datasets. Some scholars argue that BA is indivisible and hard to fit for parallel computing \cite{zhou2020stochastic}. To solve this problem, they have turned to finding approximate methods that are more suitable for parallel computing \cite{toselli2004domain, eriksson2016consensus, natesan2017distributed, zhang2017distributed, zhu2017parallel, demmel2020distributed, zhou2020stochastic, huang2021deeplm}. On the contrary, other works still utilize the LM algorithm, and attempt to design special parallel frameworks for it \cite{wu2011multicore, maoteng2017new, ren2022megba}. The former studies often give sub-optimal results, while the latter methods, despite running fast, are still limited to median-scale datasets (about 10k images).

%-------------------------------------------------------------------------
\section{Related works}
\label{section2}
For large-scale datasets at the city- or even world-level, some works just register images and give an approximate reconstruction \cite{frahm2010building, klingner2013street,jared2015reconstructing, shen2016graph}. Heinly et al. \cite{jared2015reconstructing} reported a method capable of processing a large dataset with 100 million images, and more than one million images were reconstructed. But the global bundle adjustment is never conducted due to the large memory requirement and heavy computation complexity. Some works try to optimize the global map, the most of them apply a divide-and-conquer strategy, partitioning the global map into submaps and then solve the submaps in parallel. The connection between submaps is either alternately optimized with submaps \cite{ni2007out}, or solved with alternating direction method of multipliers (ADMM) \cite{eriksson2016consensus, natesan2017distributed, zhang2017distributed, demmel2020distributed, zhou2020stochastic}, global motion averaging \cite{govindu2004lie, hartley2013rotation, zhu2017parallel}and domain decompositions\cite{toselli2004domain, huang2021deeplm}. These methods, however, often give sub-optimal results because they are only approximations to the LM algorithm. As far as we know, the largest dataset processed by strict LM algorithm only includes about 20K images \cite{ren2022megba}. 

Another line of methods apply the traditional LM algorithm while using high-performance GPU hardware to accelerate the process. Agarwal et al. \cite{agarwal2010bundle, agarwal2011building, agarwal2012ceres} first introduced a framework suitable for bundle adjustment with large-scale datasets. The framework is then implemented and accelerated on GPU hardware \cite{wu2011multicore, maoteng2017new}. More recently, Ren et al. \cite{ren2022megba}proposed a GPU-based distributed BA library, MegBA, and largely improved the efficiency of bundle adjustment using multiple GPUs. However, these methods consume a large amount of CPU/GPU memory, which limits their application for larger datasets.

In this paper, to maintain high accuracy, we also utilize the LM algorithm. Different from the existing distributed methods, which partition the global map to small ones and conduct bundle adjustment in the submaps, we only perform a global bundle adjustment where the formation of the global RCS is actually parallelized and executed in a distributed fasion. The 3D points are firstly partitioned to different groups, each of which is submitted to a computing node along with the involved camera poses, and then sub-RCSs are generated at computing nodes. Finally, the global RCS is aggregated by summing all these sub-RCSs.

To store the large RCS, we compress it with a block-based sparse matrix compression format (BSMC) \cite{zheng2016bundle}, which fully exploits the block feature of the RCS and subsequently achieves better performance than the widely used Compressed Sparse Row (CSR) format \cite{saad1994sparskit, bell2009implementing} on both the compression ratio and accessing efficiency. Once the global RCS is formed, it is then solved by PCG, which is also implemented in parallel (multi-threads), and finally the unknowns are calculated and updated to all computing nodes for the next iteration.

The contributions of this work are as follows:

	(1) We propose a distributed method to form the global RCS, which largely improves the efficiency and scalability.

	(2) A block-based sparse compression (BSMC) format is applied to store the RCS and sub-RCS, and the compression ratio and the accessing efficiency of the compressed structure are largely improved.

	(3) The proposed pipeline is compared with state-of-the-art methods. Numerous experimental results demonstrate the superiority of our proposed method over the baselines regarding memory usage and scalability. 

    (4) For the first time, we conduct parallel bundle adjustment using the exact LM algorithm on super-large datasets with up to 10 million images (500 times that of the state-of-the-art).

\begin{figure*}
  \centering
  \includegraphics[width=0.9\linewidth]{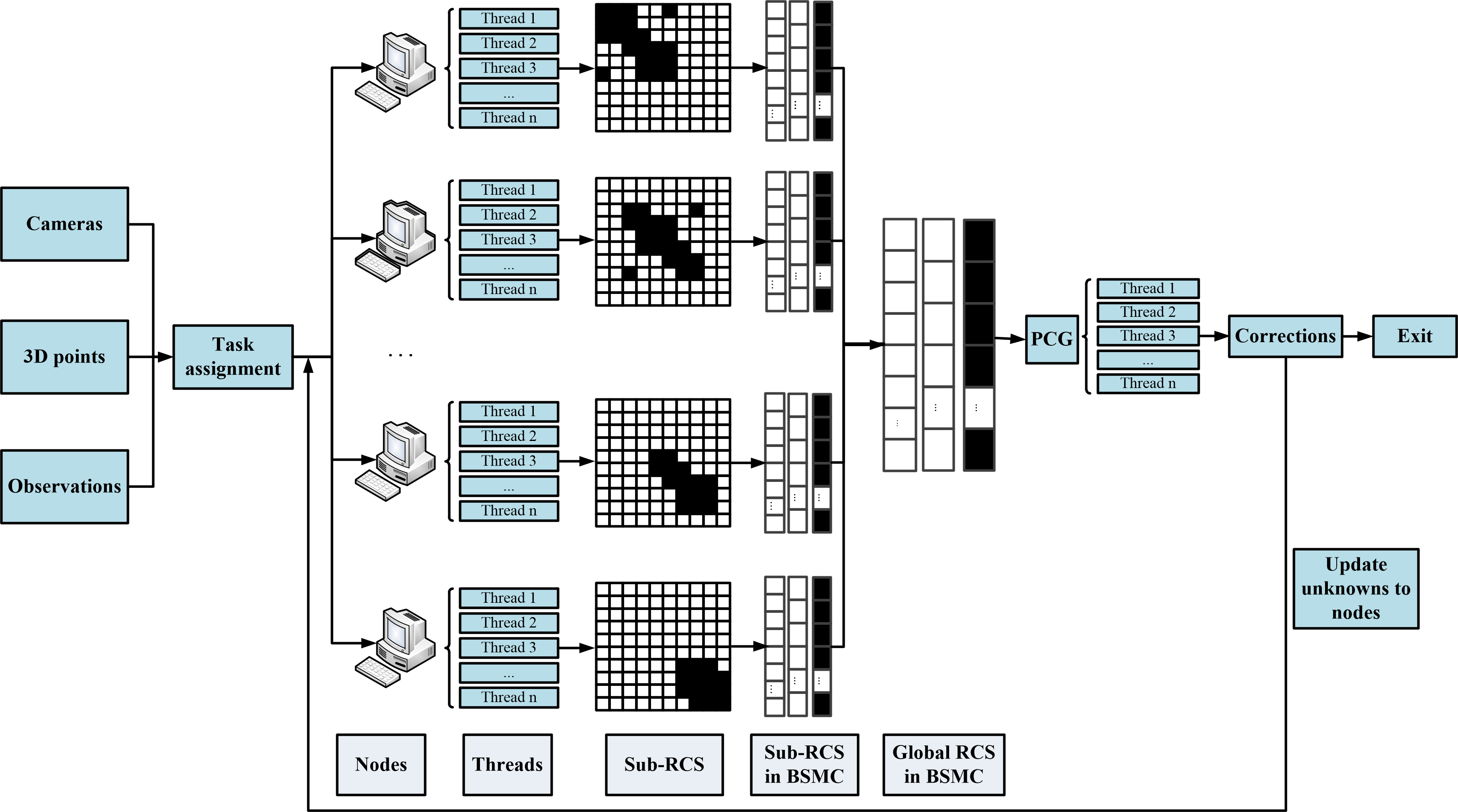}
  \caption{The pipeline of the proposed method. First, tie points are divided into groups and submitted to computing nodes along with the initial camera poses; then the sub-RCSs are generated and stored in BSMC format at the computing nodes; after that, the global RCS is obtained by aggregating all the sub-RCSs and then solved in parallel using PCG on the main node; finally, the camera unknowns are computed on the main node and updated on the computing nodes for the next iteration until the convergency is achieved.}\label{fig2}
\end{figure*}

\section{Problem formation}
\label{section3}
\subsection{The LM algorithm}
\label{section3.1}
Bundle adjustment optimizes the camera poses as well as the 3D points by minimizing the square sum of the reprojection errors, and has been widely studied for decades. First, we begin with the normal equation as follows.
\begin{equation}\label{eq1} 
(\mathbf{J}^T\mathbf{J}+\lambda \mathbf{D}^T\mathbf{D})\Delta \mathbf{x}=-\mathbf{J}^T\mathbf{e}
\end{equation}
Where $\mathbf{J}$ is the Jacobian matrix, the damping matrix $\mathbf{D}$ is usually a diagonal matrix extracted from the matrix $\mathbf{J}^T\mathbf{J}$, and the non-negative coefficient $\lambda$ controls the damping strength. Vector $\Delta \mathbf{x}$ is the updating step of unknowns, and $\mathbf{e}$ is the vector of reprojection errors.

The unknowns can be partitioned into camera part and ground-point part such that $\Delta \mathbf{x}=[\Delta \mathbf{x}_c \quad \Delta \mathbf{x}_p]$. Accordingly, the partitioning can be also performed to the Jacobian $\mathbf{J}=[ \mathbf{J}_c \quad \mathbf{J}_p]$, and damping matrix $\mathbf{D}=[ \mathbf{D}_c \quad \mathbf{D}_p]$ where the subscripts $c$ and $p$ denote the camera part and ground-point part, respectively. Then, Equation (\ref{eq1}) can be rewritten as follows

%\begin{equation}\label{eq2}
%\left[
%   \begin{array}{cc}
%     \mathbf{J}_c^T \mathbf{J}_c+\lambda \mathbf{D}_c^T \mathbf{D}_c & \mathbf{J}_c^T \mathbf{J}_p \\
%     \mathbf{J}_p^T \mathbf{J}_c & \mathbf{J}_p^T \mathbf{J}_p+\lambda \mathbf{D}_p^T \mathbf{D}_p \\
%   \end{array}
% \right]
% \left[
%   \begin{array}{cc}
%    \Delta \mathbf{x}_c  \\
%    \Delta \mathbf{x}_p \\
%   \end{array}
% \right]=
%  \left[
%   \begin{array}{cc}
%    \mathbf{J}_{c}^T\mathbf{e} \\
%    \mathbf{J}_{p}^T\mathbf{e} \\
%   \end{array}
% \right]
%\end{equation}

\begin{equation}\label{eq3}
\left[
   \begin{array}{cc}
     \mathbf{U} & \mathbf{W} \\
     \mathbf{W}^T &  \mathbf{U} \\
   \end{array}
 \right]
 \left[
   \begin{array}{cc}
    \Delta \mathbf{x}_c  \\
    \Delta \mathbf{x}_p \\
   \end{array}
 \right]=
  \left[
   \begin{array}{cc}
    \mathbf{l}_{c} \\
    \mathbf{l}_{p} \\
   \end{array}
 \right]
\end{equation}
Where $\mathbf{U}=\mathbf{J}_c^T \mathbf{J}_c+\lambda \mathbf{D}_c^T \mathbf{D}_c, \mathbf{V}=\mathbf{J}_p^T \mathbf{J}_p+\lambda \mathbf{D}_p^T \mathbf{D}_p, 
\mathbf{W}=\mathbf{J}_c^T \mathbf{J}_p, \mathbf{l}_c=-\mathbf{J}_c^T \mathbf{e}$, and $\mathbf{l}_p=-\mathbf{J}_p^T \mathbf{e}$.

Matrices $\mathbf{U}$ and $\mathbf{V}$ are block diagonal, and matrix $\mathbf{W}$ is a block-based sparse matrix. A block-wise Gauss elimination method is usually applied to eliminate the ground-point unknowns. Then, the number of unknowns in the adjustment is largely decreased because the number of cameras is usually much smaller than the number of ground points. After the elimination of ground points, we obtain:

\begin{equation}\label{eq4}
  (\mathbf{U}-\mathbf{W}\mathbf{V}^{-1} \mathbf{W}^T )\Delta \mathbf{x}_c=\mathbf{l}_c-\mathbf{W}\mathbf{V}^{-1} \mathbf{l}_p
\end{equation}
\begin{equation}\label{eq5}
  \mathbf{V}\Delta \mathbf{x}_p=\mathbf{l}_p-\mathbf{W}^T \Delta\mathbf{x}_c
\end{equation}

	The camera unknown vector $\Delta \mathbf{x}_c$ is first solved according to Equation (\ref{eq4}), and then the ground-point unknown vector $\Delta \mathbf{x}_p$ can be substituted according to Equation (\ref{eq5}) or triangulated by camera poses. This is the so-called Schur complement trick. Equation (\ref{eq4}) is the reduced camera system (RCS). In the following context, we refer the RCS to the matrix $\mathbf{U}-\mathbf{W}\mathbf{V}^{-1} \mathbf{W}^T$ for simplicity.

\subsection{Preconditioned conjugate gradient}
\label{section3.1}
The unknowns, although largely reduced by Shur complement, are still too much for direct inversion or Cholesky decomposition of the RCS. The conjugate gradient (CG) method is preferred. It is an iterative method to solve linear systems, whereby the direct inversion of the large matrix is avoided and only a matrix-vector product needs to be iteratively computed. In Equation (\ref{eq4}), let $\mathbf{R}=\mathbf{U}-\mathbf{WV}^{-1} \mathbf{W}^T, \mathbf{y}=\Delta \mathbf{x}_c$, and $\mathbf{b}= \mathbf{l}_c-\mathbf{WV}^{-1} \mathbf{l}_p$, and then it can be rewritten as follows

\begin{equation}\label{eq6}
\mathbf{R}\mathbf{y}=\mathbf{b}
\end{equation}

The above equation can be solved by CG. In general, a preconditioner $\mathbf{M}^{-1}$ is applied to decrease the CG iteration times. The main task is now shifted to solving Equation (\ref{eq7}). 

\begin{equation}\label{eq7}
\mathbf{M}^{-1}\mathbf{R}\mathbf{y}=\mathbf{M}^{-1}\mathbf{b}
\end{equation}

This is the so-called preconditioned conjugate gradient (PCG). In this paper, a block diagonal preconditioner \cite{agarwal2010bundle} is selected to conduct PCG. The detailed procedures for PCG can be found in numerous references; therefore, we omitted them here. The most time-consuming step in the PCG is a matrix-vector product. The others are only vector-vector products. Thus the parallel manipulation in Section \ref{section6.2} is only applied for the matrix-vector product
%The main procedures of solving equation (7) with PCG are listed as below:\\\\
%Set initial parameters: $\mathbf{y}_0=\mathbf{0}, \mathbf{r}_0=\mathbf{b}-\mathbf{R}\mathbf{y}_0, \mathbf{d}_0=\mathbf{M}^{-1} \mathbf{r}_0, k=0$;\\
%Do\\ 
%$\alpha_k=(\mathbf{r}_k^T \mathbf{M}^{-1} \mathbf{r}_k)/(\mathbf{d}_k^T \mathbf{Rd}_k )$\\
%$\mathbf{y}_{k+1}=\mathbf{y}_k+\alpha_k \mathbf{d}_k$\\
%$\mathbf{r}_{k+1}=\mathbf{r}_k-\alpha_k \mathbf{Rd}_k$\\
%$\eta_k=\|\mathbf{r}_0 \|_2/\|\mathbf{r}_k \|_2 $\\
%$\beta_k=(\mathbf{r}_{k+1}^T \mathbf{M}^{-1} \mathbf{r}_{k+1})/(\mathbf{r}_k^T \mathbf{M}^{-1} \mathbf{r}_k )$\\
%$\mathbf{d}_{k+1}=\mathbf{M}^{-1} \mathbf{r}_{k+1}+\beta_k \mathbf{d}_k$\\
%$k=k+1$\\
%While ( $\eta_k>T$ )\\

%Where $\mathbf{r}_k$ is the residual vector, $\mathbf{d}_k$ is the direction vector, $\alpha_k$ and $\beta_k$ are two coefficients. The subscript $k$ is the iteration number. In this paper, the threshold value $T$ is set to $0.1$ according to (Agarwal et al., 2010).

\section{Block-based sparse matrix compression}
\label{section4}
The structure of an RCS is shown in Figure \ref{fig4} (top). The non-zero blocks are determined by common points between cameras. If there are common points between camera $i$ and $j$, then the block ($i$, $j$) is a non-zero block that consists of $c_i\times c_j$ elements, where $c_i$ and $c_j$ are the number of unknowns for camera $i$ and $j$, respectively. Generally, the RCS is sparse; thus, it can be compressed by abandoning the zero elements. To this end, most of methods and libraries adopt the Compressed Sparse Row (CSR), which is a widely used sparse matrix compression format \cite{saad1994sparskit, bell2009implementing}. 

\begin{figure}
  \centering
  \includegraphics[width=0.9\linewidth]{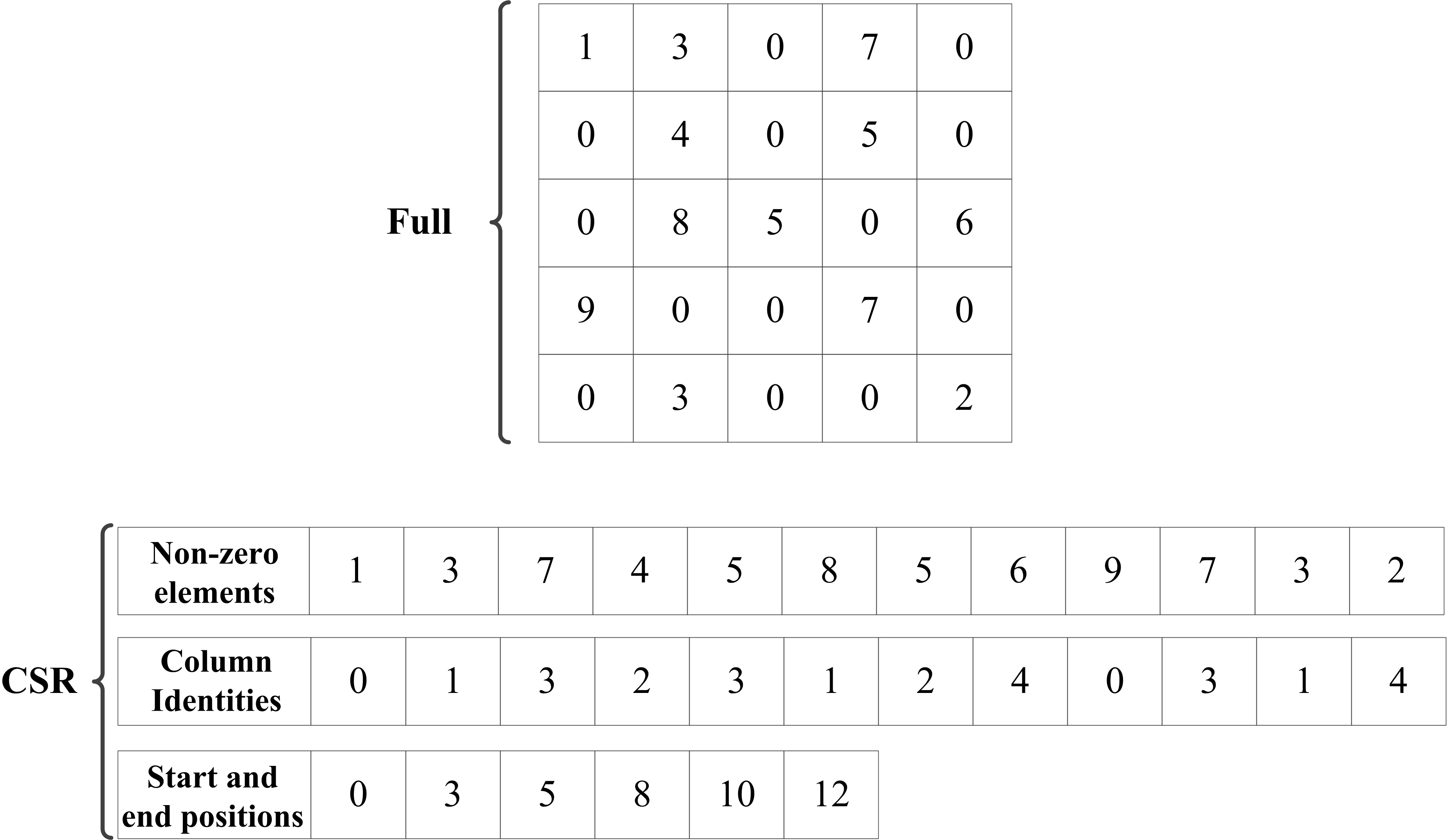}
  \caption{A full matrix (top) and the corresponding compressed structure (bottom) with CSR. Find more details in the text.}\label{fig3}
\end{figure}

\begin{figure}
  \centering
  \includegraphics[width=0.9\linewidth]{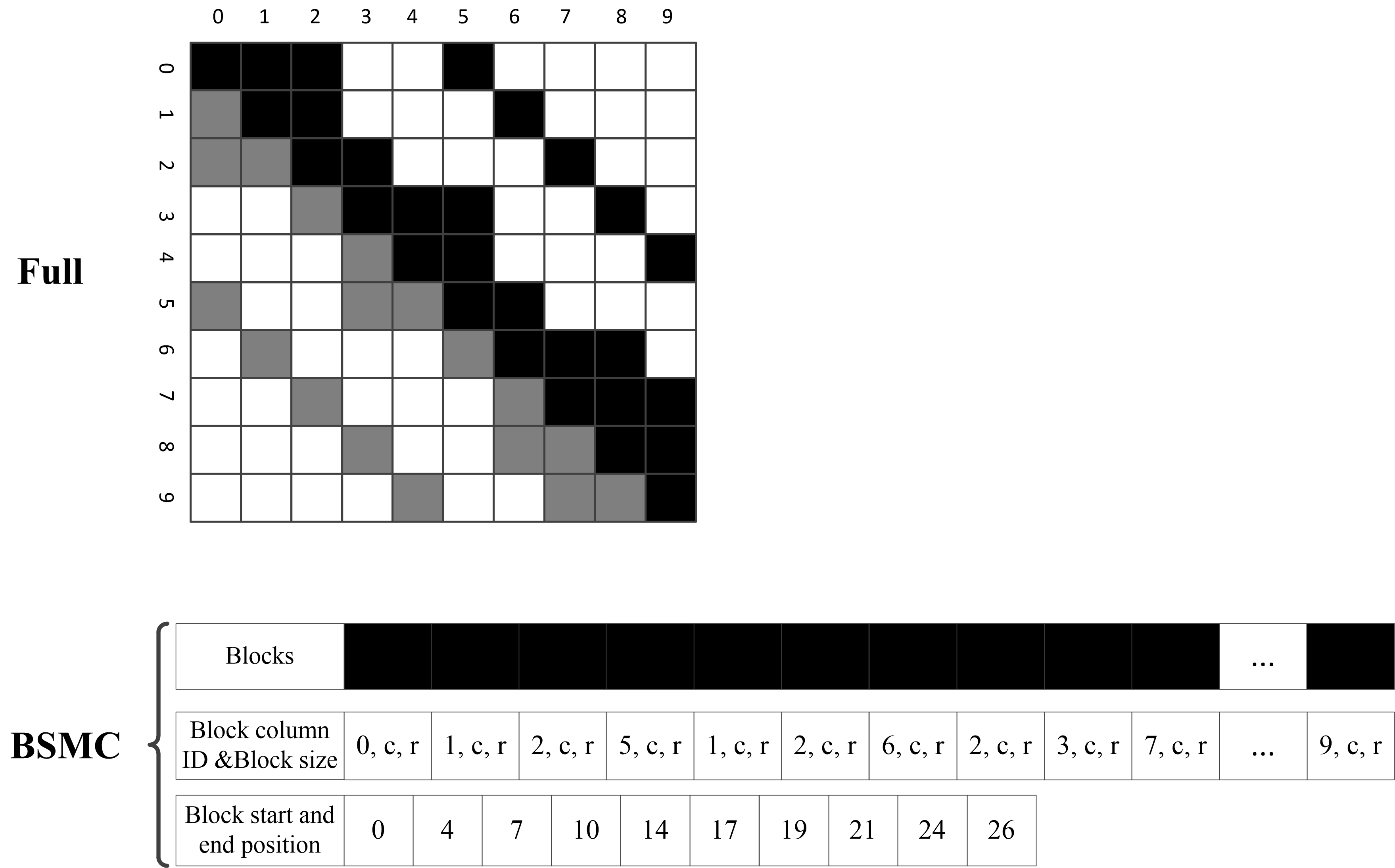}
  \caption{The structure of an RCS with ten cameras in full matrix (top) and in BSMC (bottom). Find more details in the text.}\label{fig4}
\end{figure}

The structure of CSR format consists of three arrays, the first array stores the non-zero elements, the second array stores the corresponding column identities, and the third stores the start column identities for each row (Figure \ref{fig3} (bottom)). The sizes of the first and second arrays are the number of non-zero elements in the matrix, and the size of the third array is the row number of the matrix. However, this format doesn’t take the advantage of the RCS's block characteristic (Figure \ref{fig4} (top)). The Block-based Sparse Matrix Compression (BSMC) format is more suitable \cite{zheng2016bundle}. It takes the block as the minimal storage unit. Only the non-zero blocks in the upper triangle part, as well as their positions (column and row IDs) and sizes (width and height), are stored (Figure \ref{fig4} (bottom)).


The BSMC format is more efficient than the CSR format as demonstrated in Figrue \ref{fig5} and Table \ref{tab1}. A good compression format should meet two requirements: the compression rate (the ratio of the compressed size and full size of the sparse matrix) should be as small as possible, and the accessing efficiency (decompression speed) should be as fast as possible. In this section, the BSMC is fully compared with the CSR format regarding these criteria.

\subsection{Compression rate}
\label{section4.1}
The compression rate is defined as the ratio of the compressed size to the full size. The smaller the compression ratio is, the better the compressed format performs. Given a sparse matrix with $n$ cameras, each of which contains $c$ unknowns, and the sparsity (the ratio of the non-zero blocks and the total blocks) is $\alpha$, the total number of non-zero blocks is $\alpha n^2$; thus, the number of non-zero blocks in the upper triangle part is $\alpha n^2-(\alpha n^2-n)/2=(\alpha n^2+n)/2$. Generally, the data type for elements is double (8 bytes), and for the identities it is integer (4 bytes). Furthermore, the BSMC only stores the upper triangle part of the RCS. As such, the memory requirements of different compression formats can be calculated as follows:
\begin{equation}\label{eq8}
Y_C=\alpha n^2*c^2*8+\alpha n^2*c^2*4+n*c*4
\end{equation}
\begin{equation}\label{eq9}
Y_B=(\alpha n^2+n)/2*c^2*8+(\alpha n^2+n)/2*3*4+n*4
\end{equation}
\begin{equation}\label{eq10}
Y_C/Y_B =(6\alpha nc^2+2c)/(2\alpha nc^2+2c^2+3\alpha +5)
\end{equation}
Where $Y_C$ and $Y_B$ are the memory requirement for CSR and BSMC, respectively.

\begin{figure}
  \centering
  \includegraphics[width=0.8\linewidth]{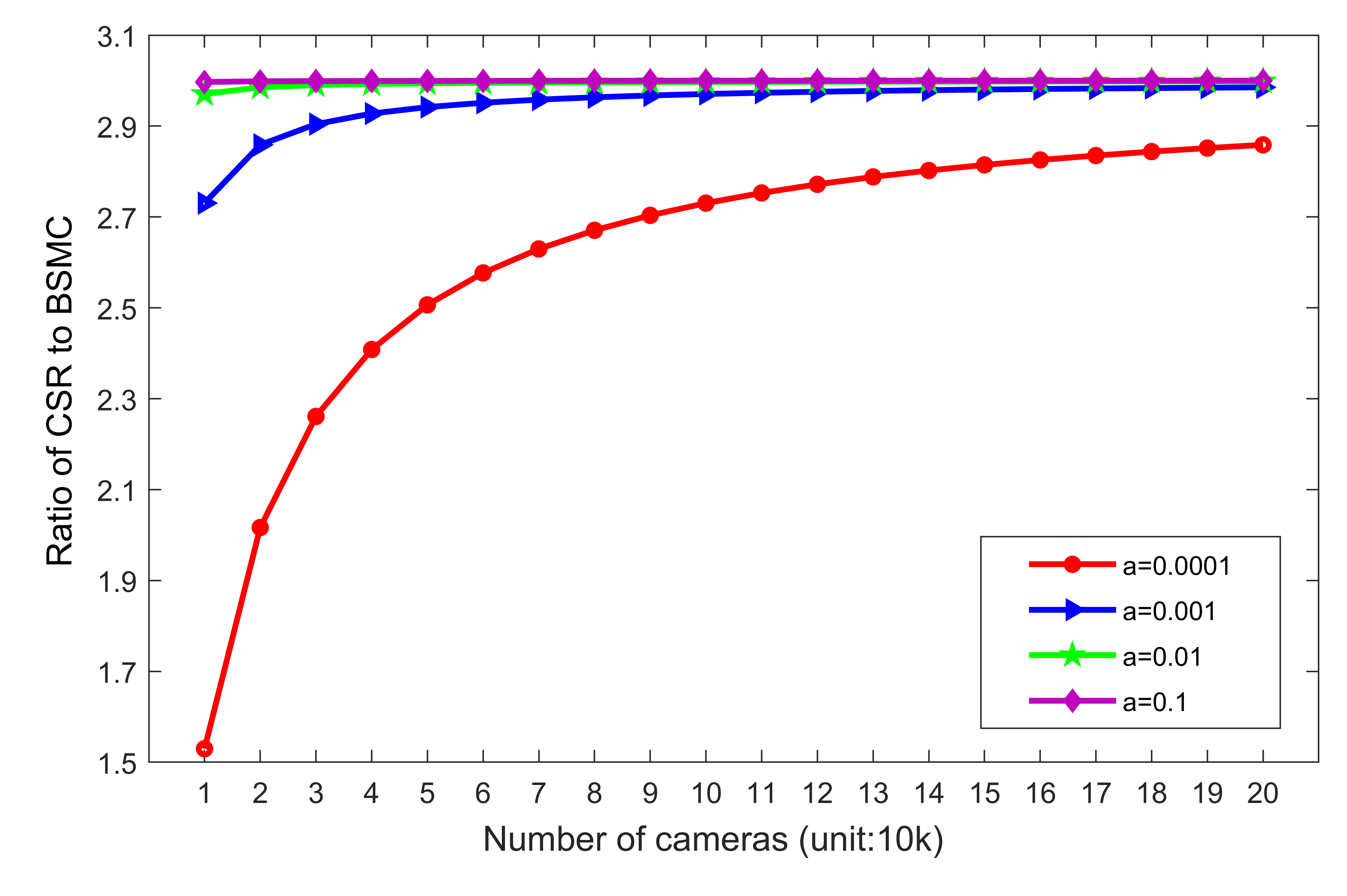}
  \caption{The relationship between the ratio of CSR to BSMC and the number of cameras for different sparsity $a$.}\label{fig5}
\end{figure}

To compare the memory requirements of the two formats, the ratio of $Y_C$ and $Y_B$ is computed by Equation (\ref{eq10}). To be specific, the number of camera parameters $c$ is set to 11 (6 for exterior and 5 for interior). We plotted the relationship between the ratio and the number of cameras ranging from 10k to 200k with different sparsities (Figure \ref{fig5}). As can be seen, the BSMC format is 1.5 to 3 times more efficient than the CSR format, especially for large-scale datasets.
\subsection{Accessing efficiency}
\label{section4.2}
To evaluate the accessing efficiency of a compression format, we consider two operations. (1) Given an element or a block in the compressed structure, find its column and row identities in the full matrix; (2) Given the column and row identities of an element or a block in the full matrix, find the corresponding block in the compressed structure.

\begin{table}
\begin{center}
  \begin{tabular}{|l|c|c|}
  \hline
  Format & First Operation & Second Operation \\
  \hline\hline
  CSR & 1 or $\log_2nc$ & $\sum^c_{i=0}\log_2cs_i$ \\
  \hline
  BSMC & 1 or $\log_2c$ & $\log_2s_i$ \\
  \hline
  \end{tabular}
\end{center}
\caption{Accessing efficiency of the BSMC and CSR formats}\label{tab1}
\end{table}
Table \ref{tab1} shows the searching times of these two operations for the BSMC and CSR formats. Readers can find more details in the Supplementary Material. As shown in Table \ref{tab1}, if the elements or blocks in the compressed structure are accessed continuously, the first operation will be immediately performed for the two formats. In the real case, for matrix-vector product, the elements or blocks in the compressed structure are continuously accessed. However, for matrix updating or aggregation, the second operation is frequently applied. The accessing efficiency of BSMC for both the first and second operations is better than the CSR; thus, it’s strongly recommended for the compression of block-based sparse matrices, such as the Hessian and RCS. 

\section{Formation of the RCS}
\label{section5}
The formation of the RCS is a major step in BA, dominating the most part of the computation complexity. In this section, we examine the details of the formation of an RCS and try to parallelize it. We find that the formation of the RCS can be performed point by point as shown in Figure \ref{fig16}. The proof of this is omitted here due to the length limitations. Readers can find the details in the Supplementary Materials.

The formation of an RCS can be summarized as follows:

(1) For each 3D point $k$, compute its Jacobian and Hessian, and then conduct the Shur complement to obtain a sub-RCS;

(2) Update the RCS with the sub-RCS;

(3) Stop until all the 3D points are completed.

	According to the above procedures, the formation of the RCS can be divided into parallel tasks, with each consisting of a group of 3D points. Those tasks are independent to each other; therefore, they can be executed in a distributed way.

\begin{figure}[t]
\begin{center}
\begin{minipage}{0.32\linewidth}
\includegraphics[width=1.0\linewidth]{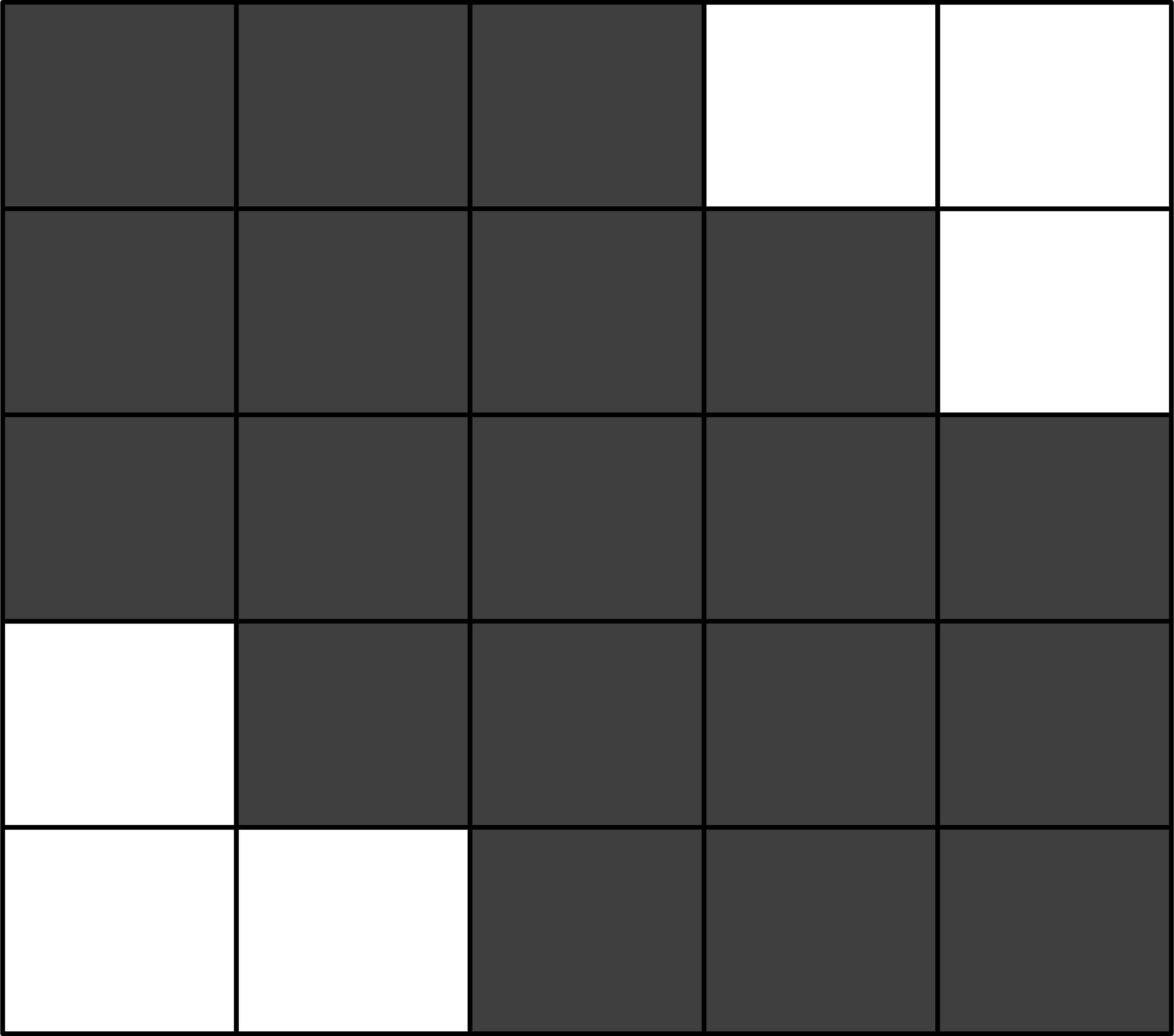}
\end{minipage}
\begin{minipage}{0.32\linewidth}
\includegraphics[width=1.0\linewidth]{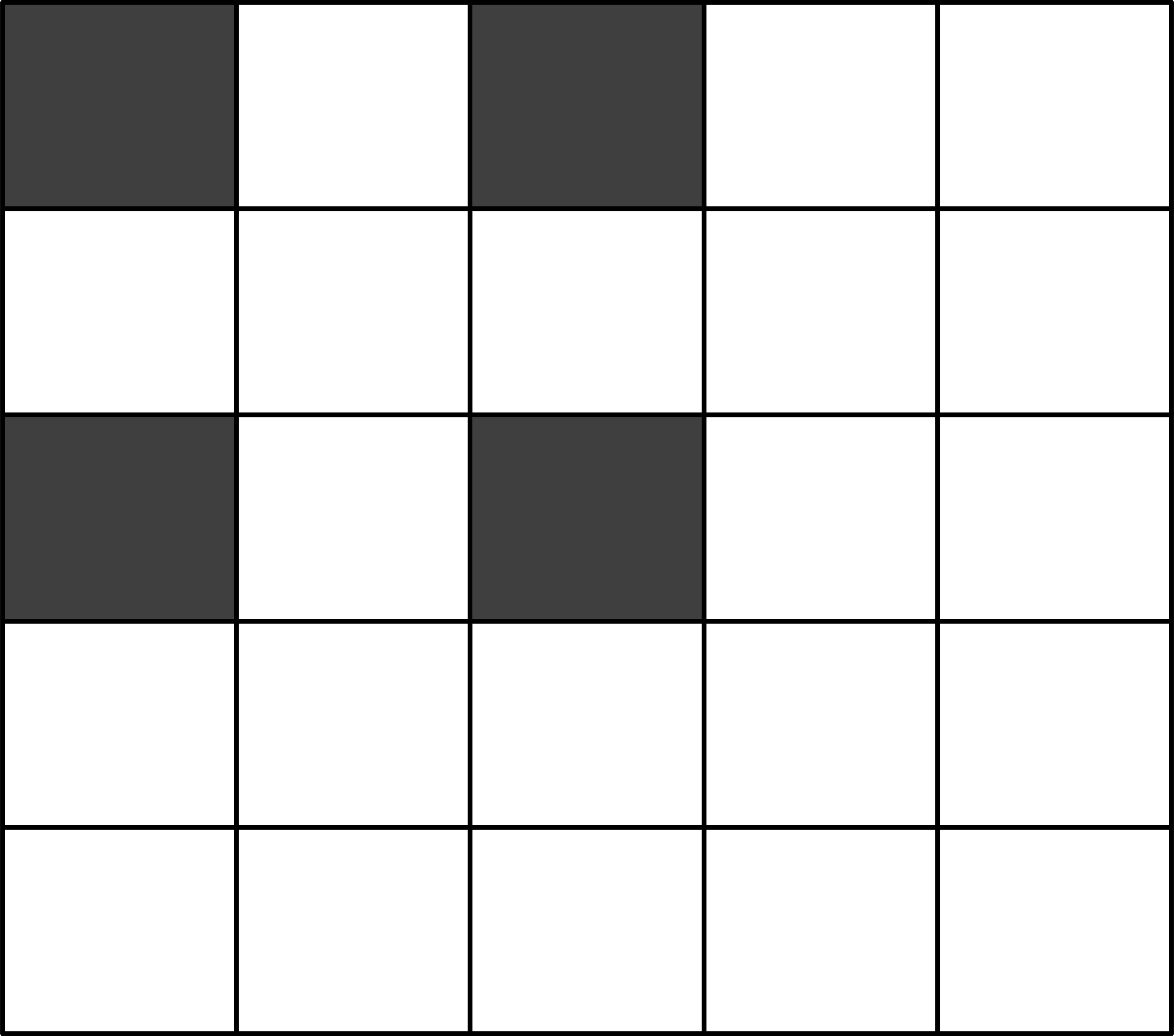}
\end{minipage}
\begin{minipage}{0.32\linewidth}
\includegraphics[width=1.0\linewidth]{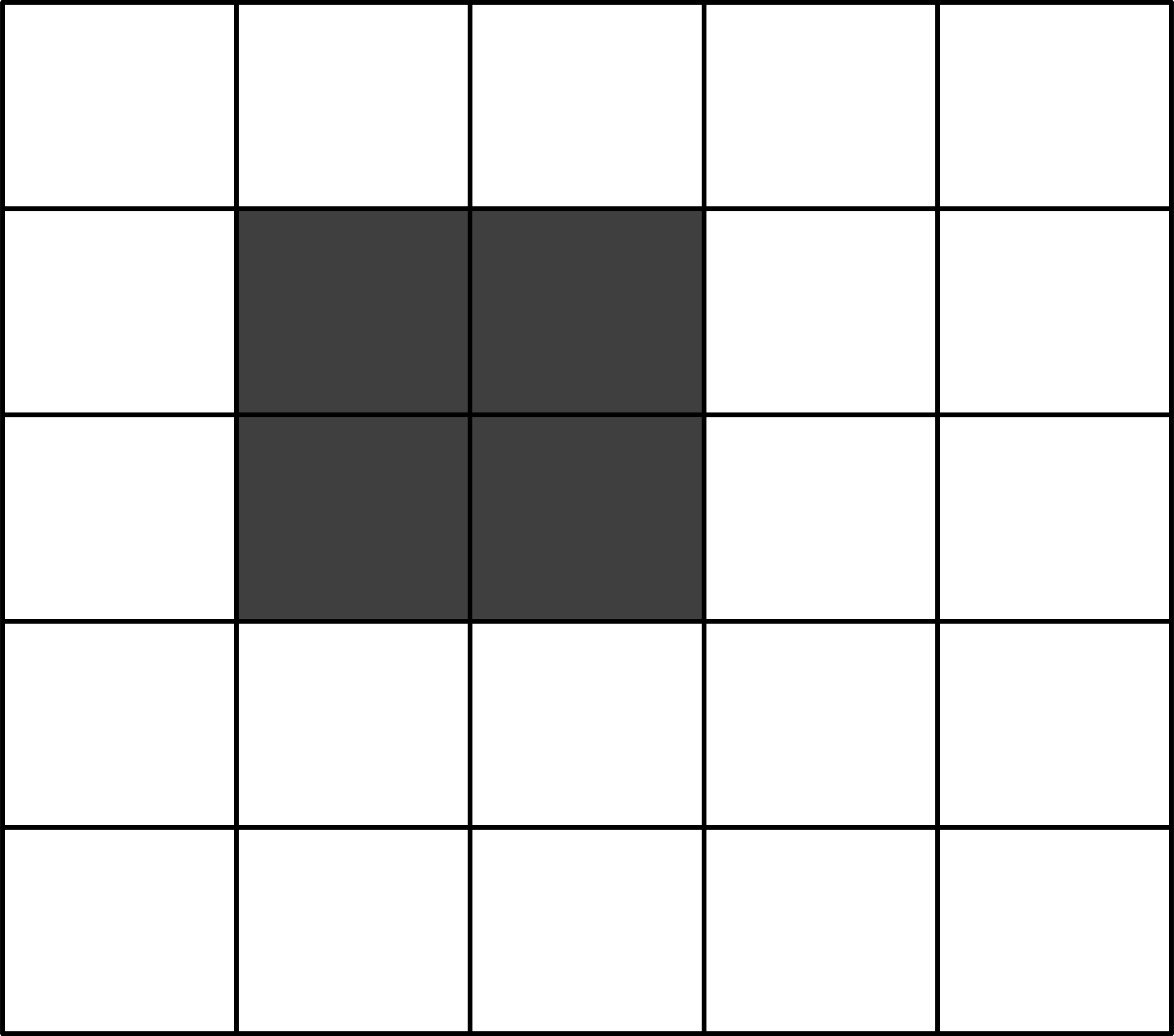}
\end{minipage}
\end{center}
   \caption{The formation of the RCS. A global RCS (left), consisting of 5 cameras, is aggregated by sub-RCSs generated by all 3D points. The right side shows two sub-RCSs. The middle is generated by a 3D point seen by camera 1 and 3, and the right is genrated by a 3D point seen by camera 2 and 3, respectively. }
\label{fig16}
\end{figure}

\section{Parallelization}
\label{section6}
The time-consuming steps in BA include the formation of the RCS, which dominates most part of the computational complexity, the computation of PCG and the triangulation of 3D points. In this paper, the formation of the RCS is executed in a distributed way, and the computation of PCG and the triangulation of 3D points are both accelerated by multi-threading. However, the parallelization of triangulating 3D points is omitted here because it's simple.
\subsection{Parallelizing the formation of the RCS}
\label{section6.1}
To parallelize the formation of the RCS, we first divide the 3D points to groups, with each submitted to a computing node along with the involved camera poses; and then generate sub-RCSs at computing nodes. Finally, we aggregate all the sub-RCSs to form a global RCS as shown in Figure \ref{fig2}. To be simple, we naively divide 3D points to groups with the same point number in this paper.

%\begin{figure}[t]
%\begin{center}
%\includegraphics[width=0.9\linewidth]{figures//figure12.png}
%\end{center}
%   \caption{The formation of the RCS in parallel. Find details in the text.}
%\label{fig12}
%\end{figure}

Note that, the sub-RCSs are first aggregated at the computing nodes to form a large sub-RCS, and then the large sub-RCS are aggregated to form a global RCS. We refer the large sub-RCS to sub-RCS for simplicity in the remainder of this paper. The sub-RCSs are directly stored in the main data center, which is connected with all computer nodes. The camera identities are reordered in the sub-RCS for efficient memory usage because they only involve a small part of the whole map as shown in Figure \ref{fig2}. The sub-RCSs and global RCS are all stored in BSMC structure. The column and row IDs of the blocks in the global RCS are stored in the second array of the BSMC structure for aggregating purpose as shown in Figure \ref{fig4} (bottom). In the meantime, an extra array is added to store the local column and row IDs in the sub-RCS to update the sub-RCSs. The formation of the sub-RCS on each computing node is further parallelized by multi-threading, making the process even more efficient.
\subsection{Parallelizing the PCG}
\label{section6.2}
\begin{figure}[t]
\begin{center}
\includegraphics[width=0.8\linewidth]{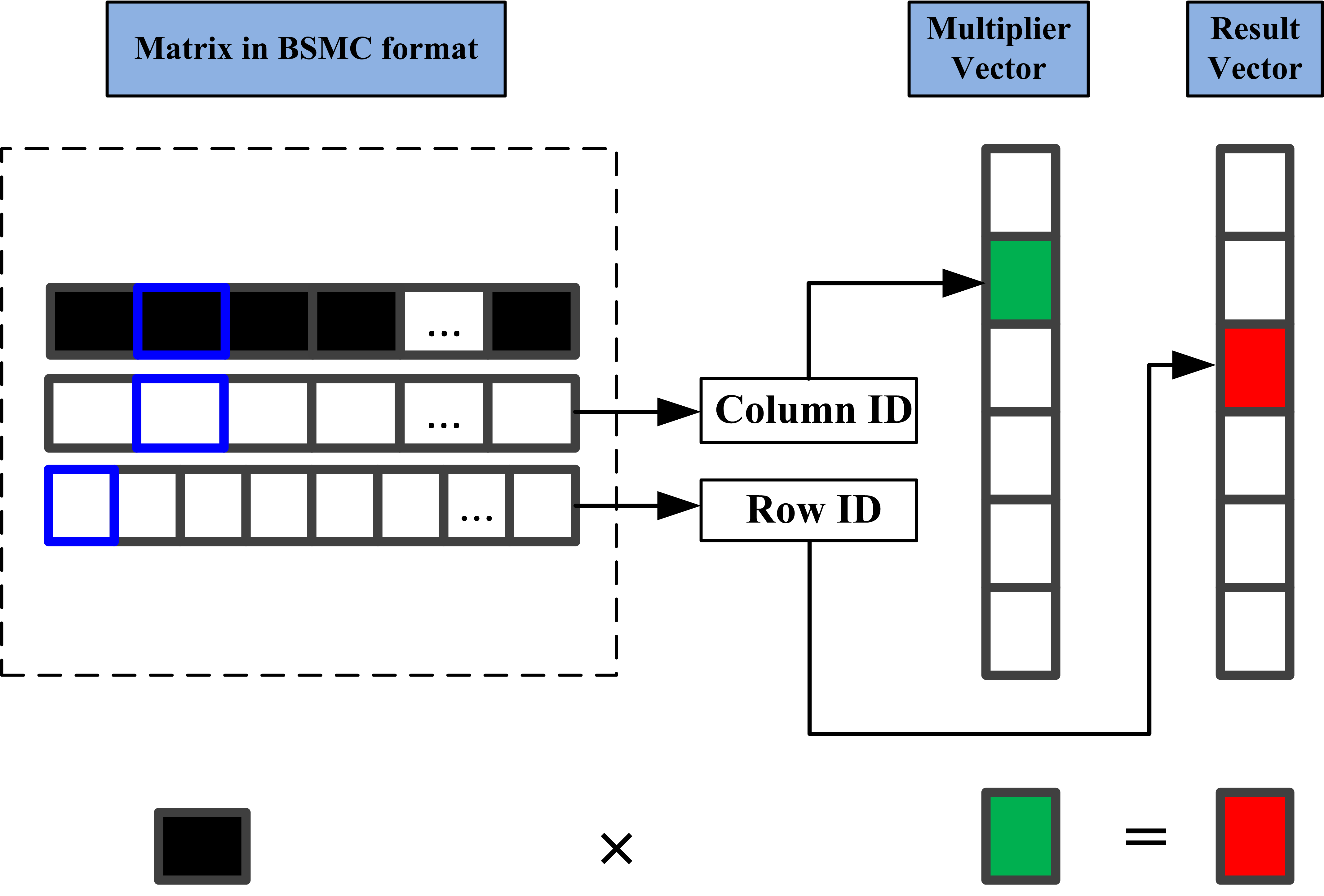}
\end{center}
   \caption{Multiplication of a matrix in BSMC format to a vector. Find details in the text.}
\label{fig13}
\end{figure}

Once the global RCS has been generated, the PCG algorithm is then applied to solve it. The most time-consuming step in the PCG is the multiplication of the RCS to the direction vector. Note that the RCS is stored in BSMC format. So, we first investigate the direct multiplication of a vector by a matrix in BSMC format. 
	As demonstrated in Figure \ref{fig4} (bottom) and Figure \ref{fig13}, the blocks are stored in the first array of the BSMC structure. We can successively read out the blocks and their column IDs in the full matrix, finding the row IDs in the full matrix by row counting according to the third array. The corresponding positions in the multiplier and result vectors can be determined by the block's column and row IDs in the full matrix, respectively. For each block, multiply it to the corresponding multiplier vector (green rectangle in Figure \ref{fig13}), and a component (red rectangle in Figure \ref{fig13}) of the result vector is generated. Finally, the result vector is obtained by aggregating all its components.

The multiplication of a BSMC-format matrix to a vector can be divided into the multiplication of blocks to the corresponding vectors as shown in Figure \ref{fig13}. Therefore, it is simple to parallelize this type of operation: First, divide the blocks into a number of groups, then submit them to multiple threads along with the corresponding multiplier vector, and finally collect all the outcome vectors and aggregate them to obtain the result vector.

%\section{Pipeline of the distributed bundle adjustment}
%The pipeline of the proposed method is show in Figure 2. The detail procedures for the whole framework are summarized as follows.

%(1) Divide landmarks into m groups;

%(2) Submit each group to a computing node along with the initial camera poses;

%(3) Compute the sub-RCS at each node with multiple threads, and store them in BSMC format;

%(4) After all tasks are completed, aggregate the sub-RCSes to form a global RCS in BSMC format;

%(5) Solve the global RCS with PCG algorithm in parallel as described in Section 7.2;

%(6) Compute and update the camera parameters;

%(7) Determine whether to stop the iteration, if yes, output the adjustment report and exit; otherwise, update the camera parameters for all the computing nodes for the next iteration, the 3D points are triangulated using the updated camera parameters on the computing nodes.

\section{Experiments}
\label{section7}
The proposed method was implemented using C++ language. To setup the parallel environment, we used ten interconnected computers to form a Local Area Network (LAN). The main program was deployed on one of the computers called the main node, and the parallel tasks were executed on other computers called computing nodes. All the data were stored in the main data center, which was connected with all nodes. The communication speed between nodes and the data center can achieve 1GB/s. The computing nodes were equipped with Inter(R) Xeon(R) CPU E5-1650 v3 @3.5GHz, 64GB RAM and Windows 10 Operating System. The main node was equiped with the same CPU and operating system, but a larger 128GB RAM. A parallel task submission and management platform provided by Mirauge3D Technology Inc. was used for parallel task assignment. The whole pipeline is also integrated in Mirauge3D, which is a commercial 3D reconstruction software available at \url{www.mirauge3d.com}. The code for BSMC is available at \url{https://github.com/MozartZheng/DistributedBA}, the public version of the whole pipeline will also be released when it is ready.

The proposed method is compared with the state-of-the-art bundle adjustment methods, such as Ceres \cite{agarwal2012ceres}, PBA \cite{wu2011multicore}, DeepLM \cite{huang2021deeplm} and MegBA \cite{ren2022megba}. These methods' experimental results on the public datasets were directly extracted from the papers of DeepLM and MegBA. To be as objective as possible, we did’t use the result of DeepLM method in its own paper; instead, we use the result provided by the paper of MegBA.

\subsection{Datasets}
\label{section7.1}
To fully evaluate the proposed method, we test it with both synthetic and real datasets. Three public datasets, Ladybug, Venice, and Final were used to test and compare the proposed method with the state-of-the-art methods. The large-scale datasets collected by UAVs were used to test the scalability of the proposed method. The image numbers of those datasets range from 21K to 1.18M. Other methods, such as PBA, DeepLM, and MegBA were not tested with these datasets because they ran out of GPU memory. For Ceres, it encountered some kind of integer overflow error for these large datasets. 

Moreover, to further test the proposed method's limit capacity, two synthetic datasets with 5 million and 10 million images are generated. To maintain simplicity, we intentionally generated the synthetic datasets with sparse features (about 300 features for each image) because the size of RCS is independent of image features. Random errors subject to normal distribution of (0.0, 1.0) pixels are added to these synthetic datasets. The statistics of the datasets are listed in Table \ref{tab2}. The sparsity is the ratio of non-zero blocks to total blocks in the RCS. The smaller the ratio, the sparser the RCS. The sparsities of the public datasets are the largest, possibly because of the large overlap. The real datasets are sparser because they are regularly collected by UAVs.
\begin{table}
\begin{center}
\begin{tabular}{|l|c|c|c|c|c|}
  \hline
  Dataset & Images & Cam. & 3D pt. & Obs. & Spa. \\
  \hline\hline
  Ladybug & 1723 & 1723 & 0.16M & 0.67M & 0.040 \\
  \hline
  Venice & 1778 & 1778 & 0.99M & 5.0M & 0.082 \\
  \hline
  Final & 13682 & 13682 & 4.4M & 29M & 0.070 \\
  \hline
  SW & 21K & 5 & 31M & 166M & 0.004 \\
  \hline
  JLP & 45K & 5 & 44M & 228M & 0.002 \\
  \hline
  YQC & 46K & 5 & 39M & 254M & 0.002 \\
  \hline
  JZ & 49K & 5 & 41M & 218M & 0.001 \\
  \hline
  XS & 62K & 95 & 55M & 267M & 0.001 \\
  \hline
  DG & 71K & 5 & 87M & 385M & 9e-4 \\
  \hline
  NM & 85K & 50 & 101M & 638M & 0.001 \\
  \hline
  MCZ & 86K & 8 & 98M & 588M & 0.001 \\
  \hline
  WQ & 91K & 60 & 93M & 535M & 8e-4 \\
  \hline
  DaPu & 133K & 10 & 8M & 61M & 4e-4 \\
  \hline
  HM & 407K & 35 & 352M & 1790M & 2e-4 \\
  \hline
  AJH & 442K & 35 & 112M & 1500M & 3e-4 \\
  \hline
  GDS & 1.18M & 114 & 1016M & 5084M & 8e-5 \\
  \hline
  Syn1 & 5M & 2 & 405M & 2839M & 3e-6 \\
  \hline
  Syn2 & 10M & 2 & 810M & 5836M & 1e-6 \\
  \hline
\end{tabular}
\end{center}
\caption{The statistics of the datasets. The term Cam. represents cameras, pt. represents points, Obs. represents observations, and Spa. represents the sparsity.}\label{tab2}
\end{table}

\subsection{Memory usage and run time}
\label{section7.2}
The memory requirements of a dataset are mainly determined by the number of tie points (3D points and corresponding image points) and the size of the RCS. For the proposed method, the tie points were divided to groups, with each submitted to a computing node. Suppose ten computing nodes are applied, the memory requirement of the proposed method for tie points are theoretically 1/10 that of the serial method (such as Ceres). Meanwhile, the proposed method uses the BSMC format to store the RCS, which is 1.5 to 3 times more efficient than other methods applying the CSR format, as discussed in Section \ref{section4}. The main node uses the most memory compared with other nodes because the global RCS is stored in the main node. Therefore, we define the main node's memory usage as that of the proposed method. 

To extensively evaluate the proposed method, we compared its memory usage, run time and accuracy with baselines using the public datasets. The results are listed in Table \ref{tab3}. As can be seen, the memory usage of the proposed method is the lowest since tie points are divided and stored in all nodes, and the memory-efficient format BSMC is applied to store the RCS. The proposed method is faster than Ceres but slower than the GPU-based methods, MegBA, DeepLM, and PBA. MegBA runs fastest but requires multiple GPUs, as well as a large amount of memory. DeepLM is faster than PBA and its memory requirement is smaller; however, it might provide sub-optimal results. Although the GPU-based methods are faster than Ceres, they use almost the same memory. Therefore, for larger datasets, MegBA, DeepLM, and PBA all run out of memory. Owing to the distributed computing manner and memory-efficient format BSMC, the proposed method is superior to all the baselines regarding memory usage.
\begin{figure}
	\begin{center}
			
		\begin{minipage}{0.99\linewidth}
			\begin{center}
				\includegraphics[width=2.6cm, height=2.3cm]{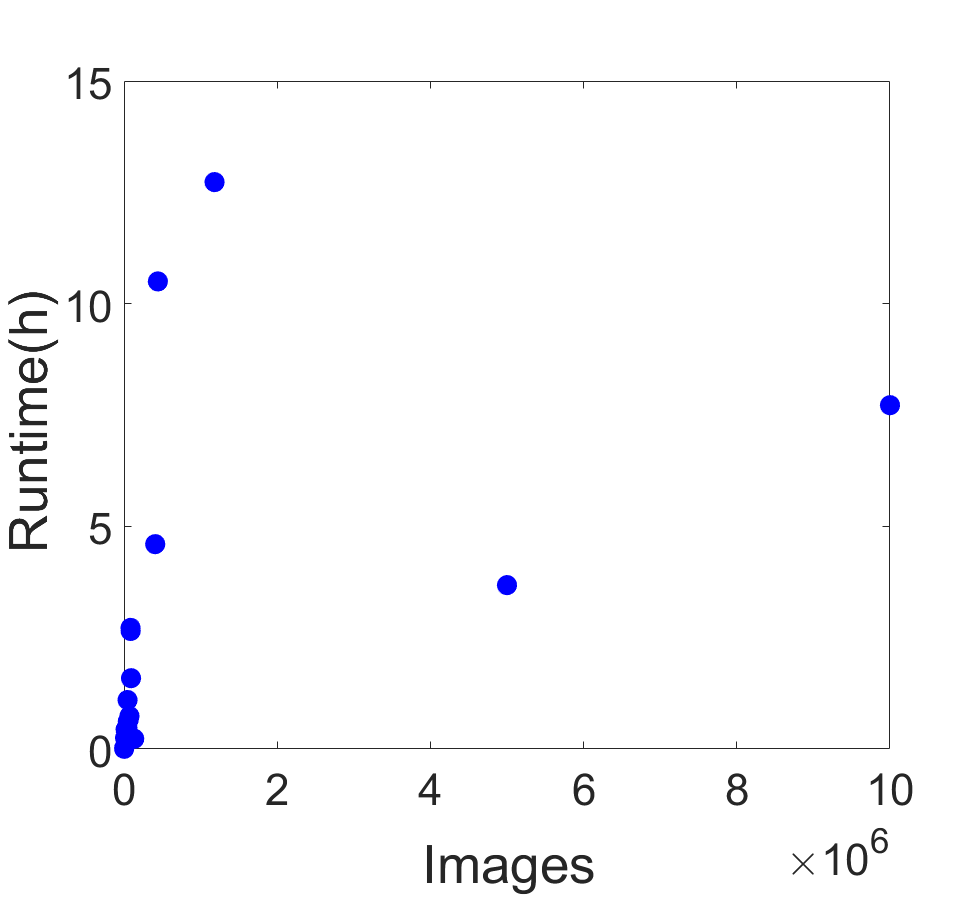}
				\includegraphics[width=2.6cm, height=2.3cm]{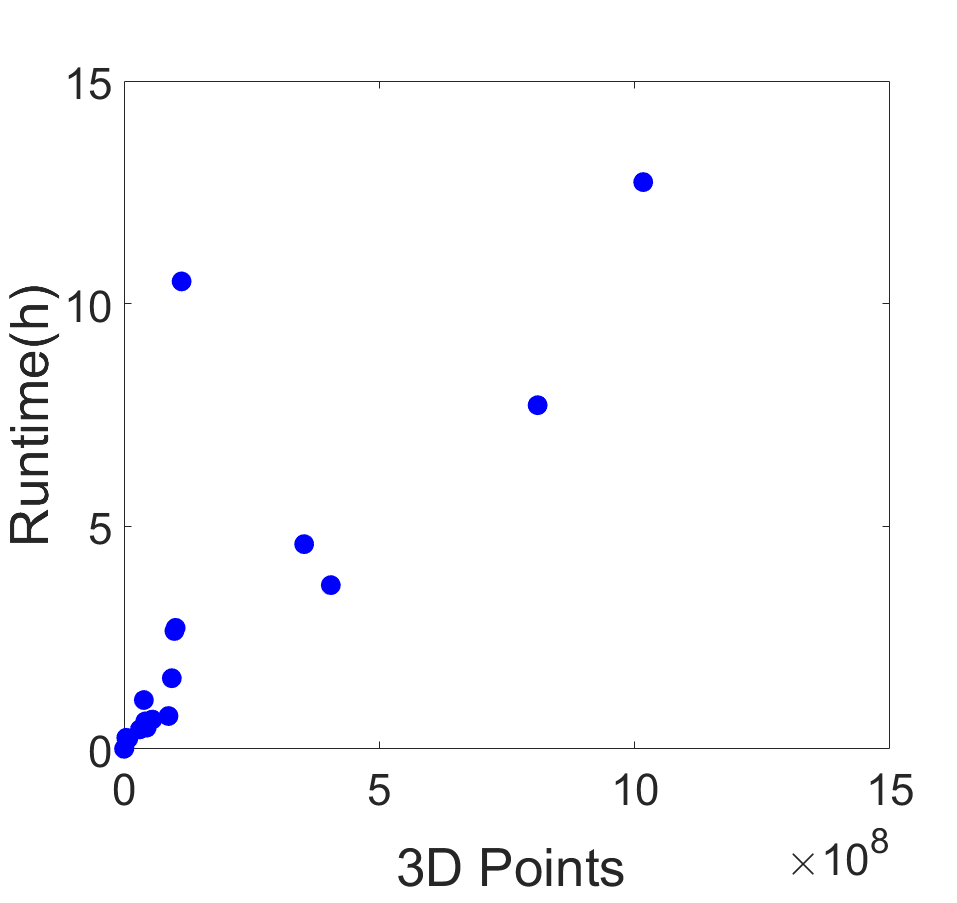}
				\includegraphics[width=2.6cm, height=2.3cm]{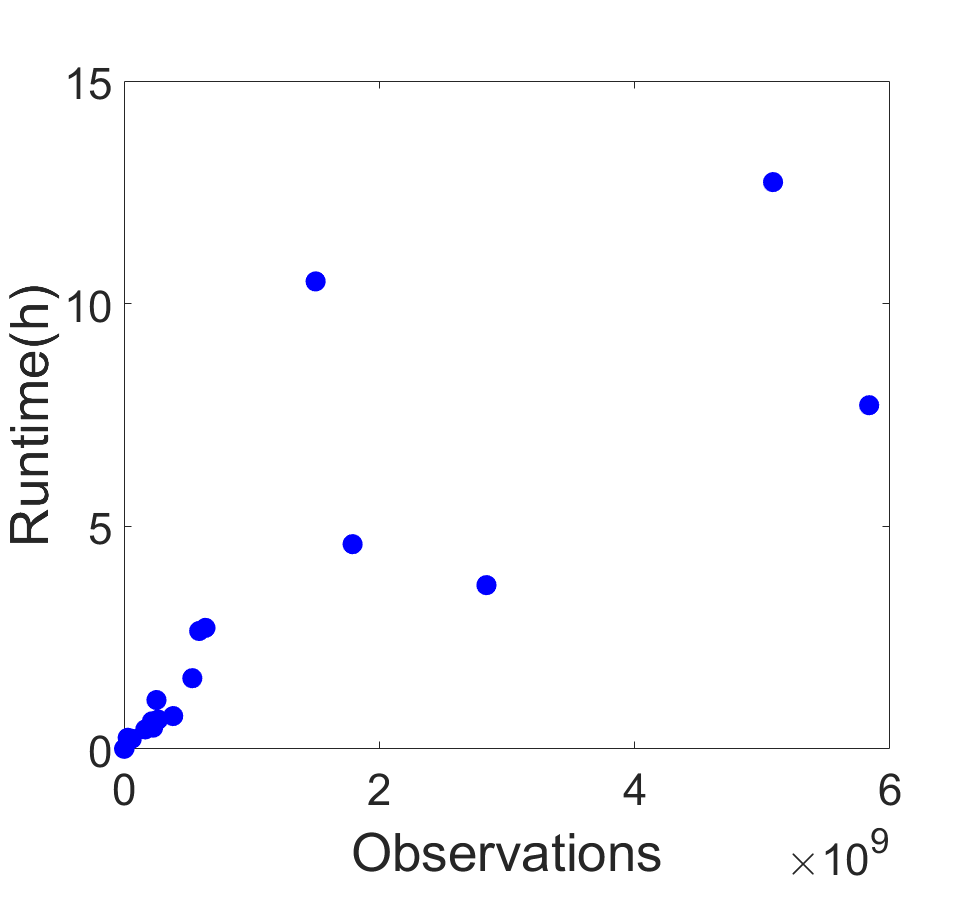}
			\end{center}
		\end{minipage}
		
	\end{center}
	\caption{The runtime with respect to various parameters.}
	\label{fig16}
\end{figure}

The super-large UAV and synthetic datasets are used to test the scalability of the proposed method. The memory usage, run time and accuracy are listed in Table \ref{tab4}. The memory requirements $M_t$ and $M_r$ are the theoretical memory required by the tie points and global RCS, respectively. The memory usage $M_u$ is the actual memory consumption for the main node. The table shows that the actual memory usage is much smaller than the memory requirement because the tie points are stored in a distributed way, whereas the main node only stores 1/10 of the tie points. For the largest dataset with 10 million images and 5.8 billion observations, the memory usage of the proposed method is only 67.48 GB, which is affordable for an advanced modern computer. 

About the communication overhead, there are three types of data need to be transferred among the nodes. First, the tie points are transferred to different computing nodes before LM iteration. Second, for each iteration, the sub-RCSs generated on the computing node should be transferred to the main node. Third, the camera poses for all images generated on the main node should be transferred back to the computing node for each iteration. As listed in Table \ref{tab5}, the time of data transmission for the largest dataset Syn2 is actually negligible compared to the total runtime which is 7.72 hours.
 \begin{table}
 	\begin{center}
 		\begin{tabular}{|l|c|c|c|}
 			\hline
 			Data & Tie points &  subRCSs & camera poses \\
 			\hline\hline
 			Syn2 & 32 & 15 & 0.2 \\
 			\hline
 		\end{tabular}
 	\end{center}
 	\caption{Data transmission time for tie points (minutes), subRCS(seconds), and camera poses (seconds) among the network for the largest dataset Syn2}\label{tab5}
 \end{table}

We plotted the run time with respect to various parameters (images, 3D points and observations) as shown in Figure \ref{fig16}. The run time is affected by various factors, the image, tie point and iteration numbers. For an instance, the run time of the synthetic dataset with 5 million images is actually about 1/4 that of the real dataset GDS which includes only 1.18 million images. The reasons could be the following: First, the tie points of the synthetic dataset are actually about half that of the GDS dataset as shown in Table \ref{tab2}; second, the iteration number is also half that of the GDS. The same result can be found in the real dataset DaPu, which has more images but less tie points.

\begin{table*}
\begin{center}
\begin{tabular}{|l|c|c|c|c|c|c|c|c|c|c|c|c|c|c|c|}
  \hline
  \multirow{2}{*}{Dataset} & \multicolumn{3}{c|}{Ceres-CG} & \multicolumn{3}{c|}{PBA} & \multicolumn{3}{c|}{DeepLM} & \multicolumn{3}{c|}{MegBA} & \multicolumn{3}{c|}{Ours}\\ \cline{2-16}
  \multirow{2}{*}{ } & M & T & Acc. & M & T & Acc. & M & T & Acc. & M & T & Acc. & M & T & Acc. \\
  \hline\hline
  Ladybug & 0.52 & 46.7 & 1.14 & 0.3 & 12.3 & 2.22 & 2.1 & 3.9 & 1.12 & 1.6 & \textbf{0.77} & 0.56 & \textbf{0.06} & 15.6 & 1.12 \\
  \hline
  Venice & 3.68 & 1992 & 0.66 & - & - & - & 6.2 & 24.4 & 0.66 & 13.6 & \textbf{11.9} & 0.33 & \textbf{0.27} & 69.7 & 0.67 \\
  \hline
  Final & 16.8 & 3897 & 1.59 & 11.9 & 340 & 3.0 & 14.89 & 149 & 1.50 & 89.7 & \textbf{22.6} & 0.75 & \textbf{4.93} & 906 & 1.24 \\
  \hline
\end{tabular}
\end{center}
\caption{Memory M(GB), run time T(s), and accuracy Acc.(pixels) of different methods.}\label{tab3}
\end{table*}

\begin{figure}[htbp]
\begin{center}
\begin{minipage}{0.99\linewidth}
\includegraphics[width=2.7cm, height=2.3cm]{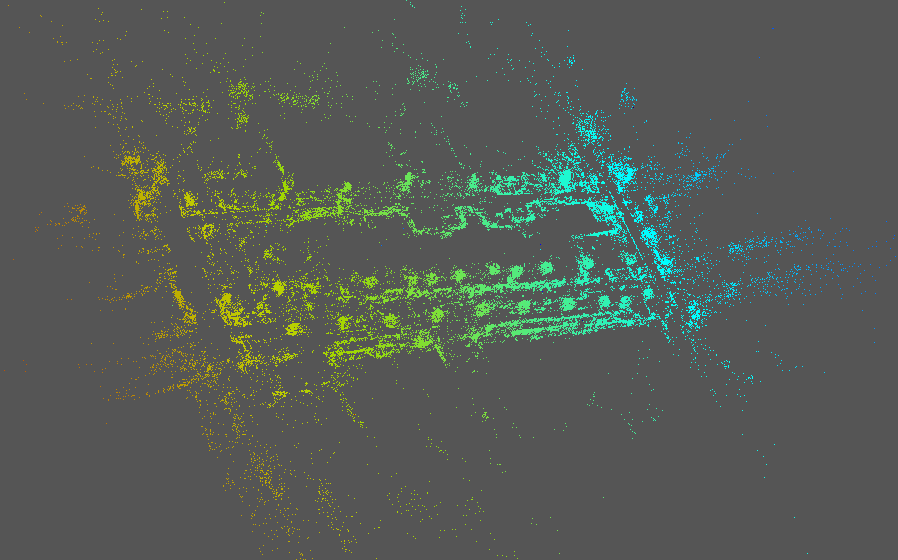}
\includegraphics[width=2.7cm, height=2.3cm]{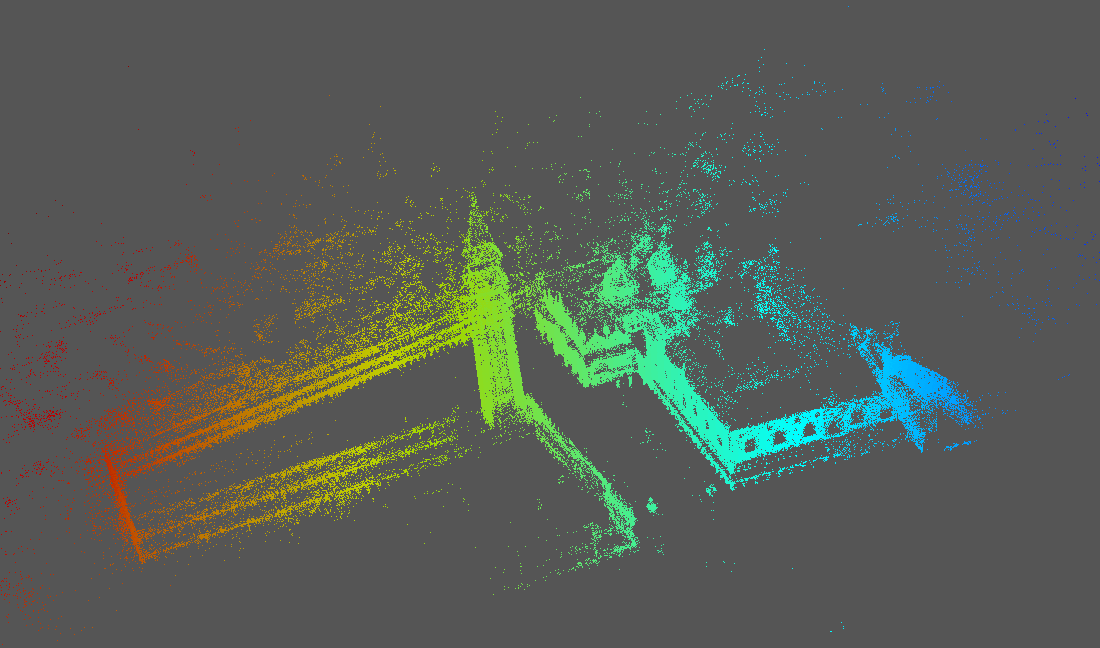}
\includegraphics[width=2.7cm, height=2.3cm]{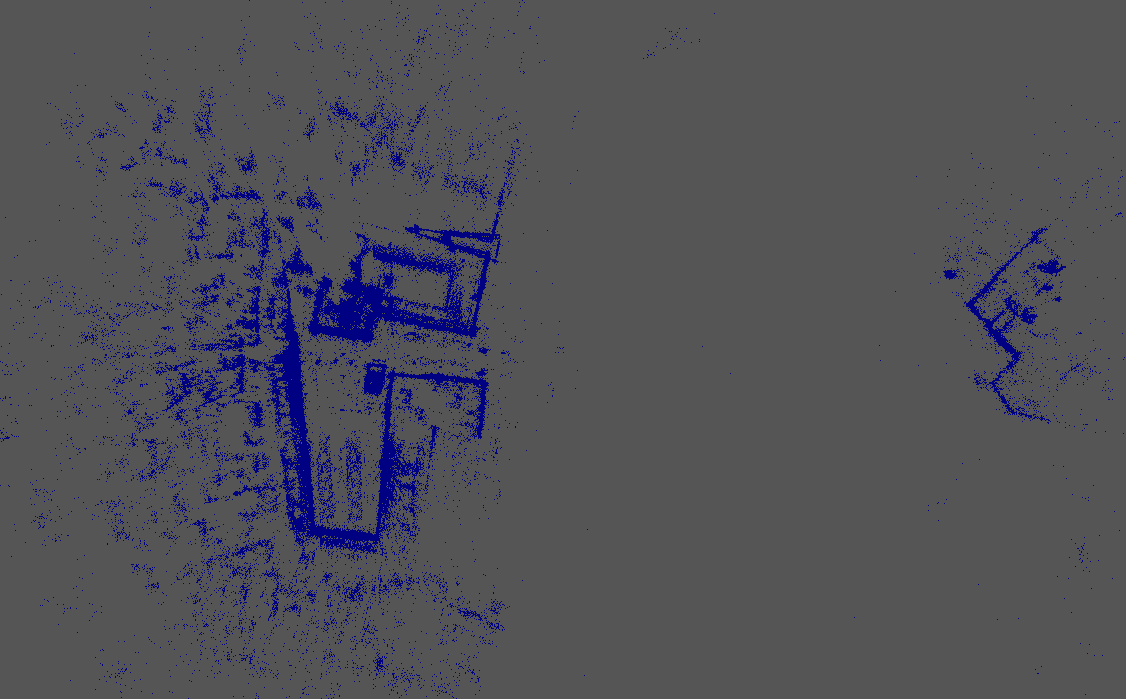}
\end{minipage}
\end{center}
   \caption{Sparse point cloud of the public datasets Ladybug, Venice and Final using the proposed method. The points in Final are all blue because there are some points far away from the main structure.}
\label{fig14}
\end{figure}

\begin{table}
  \begin{center}
  \begin{tabular}{|l|c|c|c|c|c|c|}
    \hline
Data &	$M_t$ &	$M_r$	& $M_u$ &	Iter.	&T	&Acc.\\
\hline\hline
SW	& 6.3 & 0.53 & 1.21 &	6	&0.44	&1.29\\
\hline
JLP	& 8.7 & 1.21 & 2.22  & 4	 & 0.48	 & 1.37\\
\hline
YQC	& 9.1 & 1.37 & 2.44 & 7	 & 1.10	 & 0.89\\
\hline
JZ	& 8.3 & 0.69 & 1.60  & 6	 & 0.62	 & 1.29\\
\hline
XS	& 10.5 & 0.91 & 2.06  & 6	 & 0.66	 & 1.05\\
\hline
DG	& 15.5 & 1.39 & 3.10  & 5	 & 0.74	 & 1.24\\
\hline
NM	& 23.0 & 2.46 & 5.04  & 6	 & 2.72	 & 1.22\\
\hline
MCZ	& 21.6 & 2.22 & 4.63  & 7	 & 2.65	 & 0.69\\
\hline
WQ	& 19.9 & 2.00 & 4.22  & 5	 & 1.59	 & 1.24\\
\hline
DaPu & 2.1 & 1.81 & 2.23  & 6	 & 0.23	 & 1.45\\
\hline
HM	& 69 & 9.72 & 17.73  & 6	 & 4.60	 & 1.27\\
\hline
AJH	& 46.3 & 16.18 & 	22.64  & 	6	 & 10.50	 & 1.28\\
\hline
GDS	& 197 & 28.22 & 	51.15  & 	6	 & 12.73	 & 1.23\\
\hline
Syn1 & 99.9 & 20.84 & 	33.52  & 	3	 & 3.68	 & 1.15\\
\hline
Syn2 & 204 & 41.70 & 	67.48  & 	3	 & 7.72	 & 1.17\\
    \hline
  \end{tabular}
  \end{center}
  \caption{Memory(GB), run time T(h) and accuracy Acc.(pixels) of the proposed method for super-large datasets. The terms $M_t$ and $M_r$ show the theoretical memory that required by the tie points and the global RCS respectively. The memory usage $M_u$ is the actual memory consumption on the main node. The term Iter. represents the iteration times.}\label{tab4}
\end{table}

\begin{figure}[htbp]
\begin{center}
\begin{minipage}{0.99\linewidth}
\includegraphics[width=4.67cm, height=2.5cm]{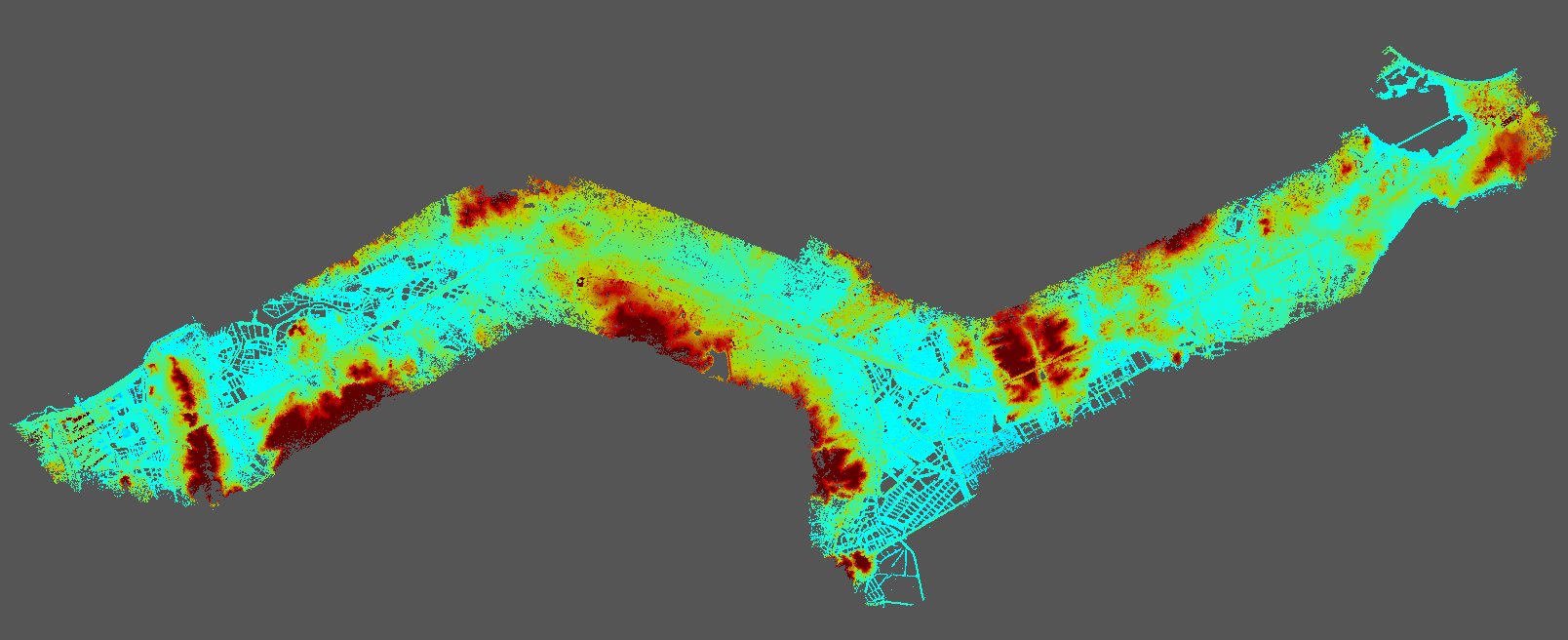}
\includegraphics[width=3.5cm, height=2.5cm]{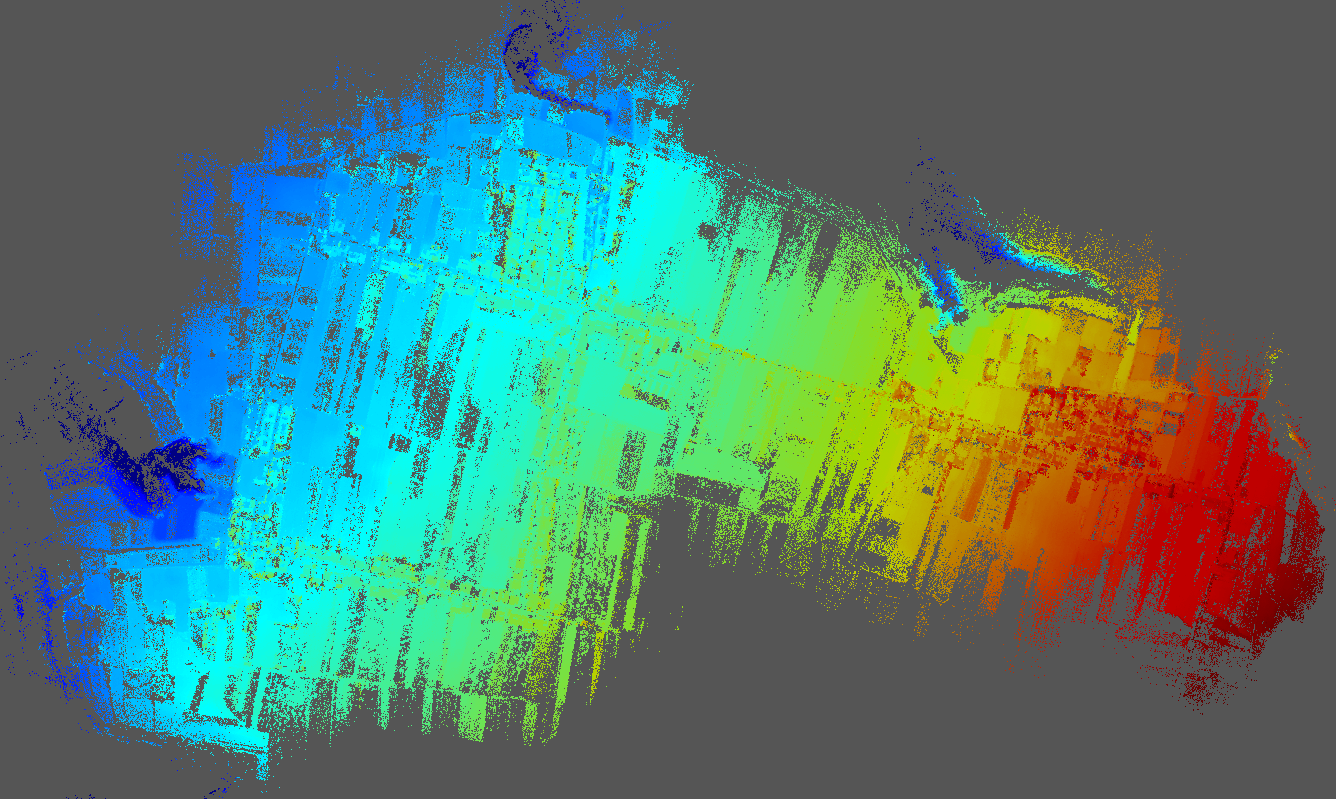}
\end{minipage}
\begin{minipage}{0.99\linewidth}
\includegraphics[width=2.7cm, height=2.5cm]{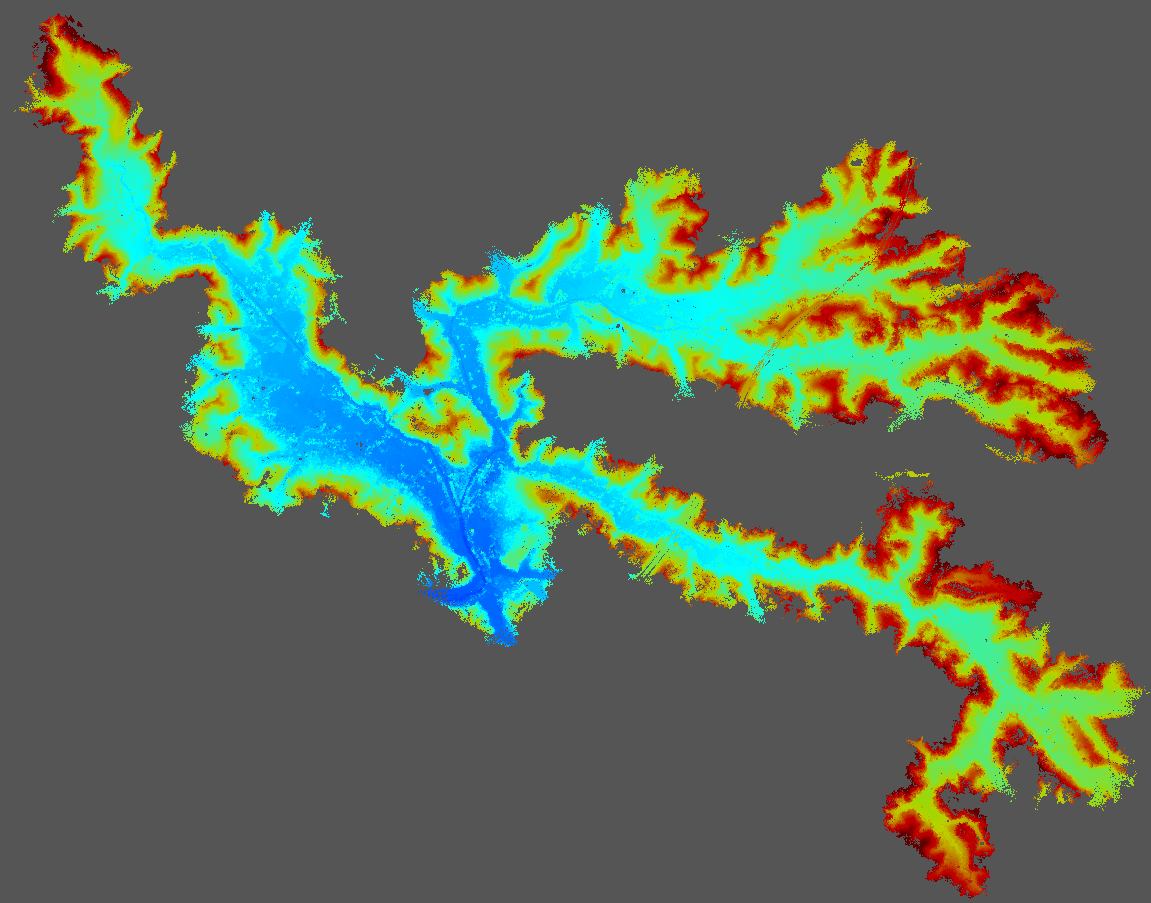}
\includegraphics[width=2.7cm, height=2.5cm]{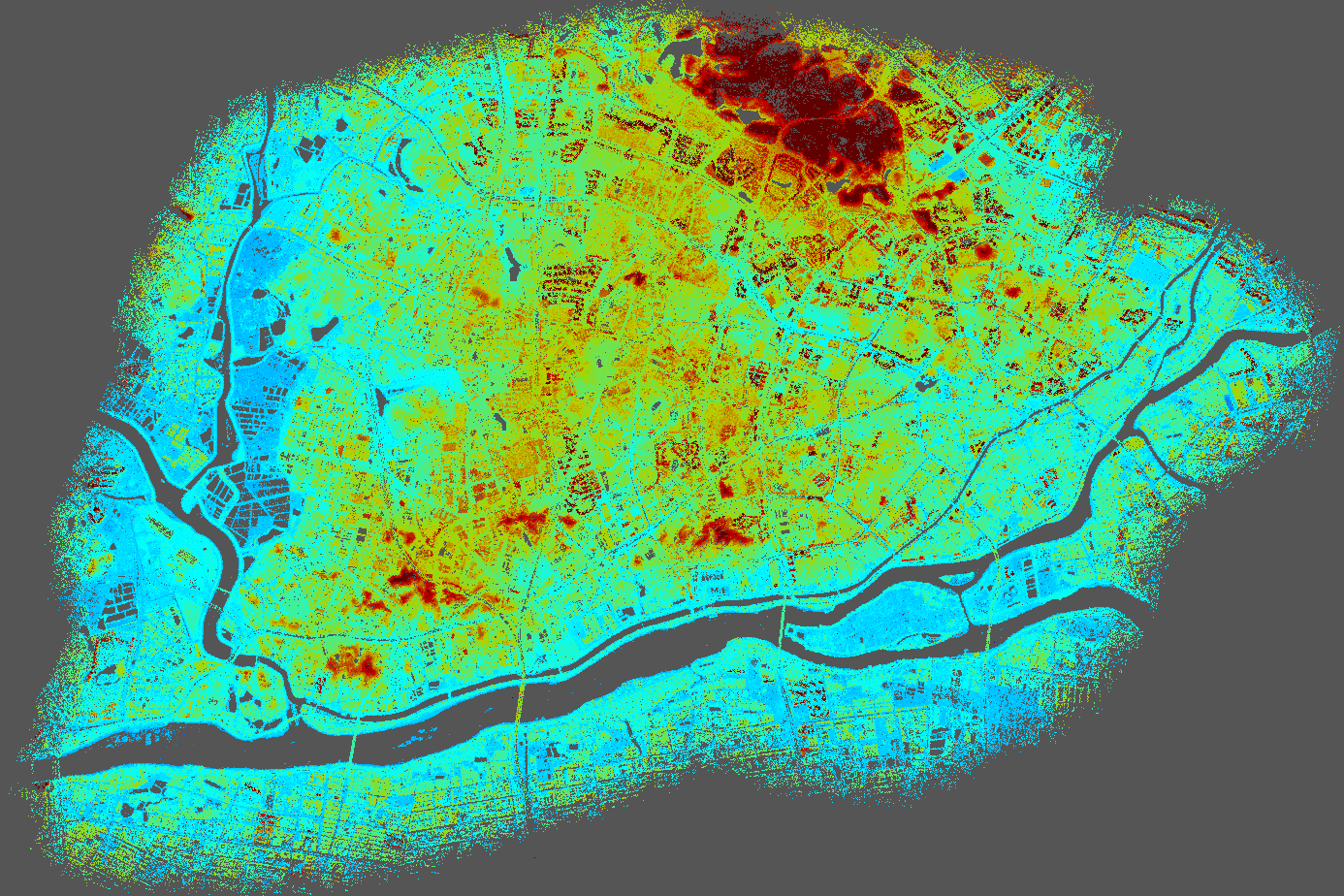}
\includegraphics[width=2.7cm, height=2.5cm]{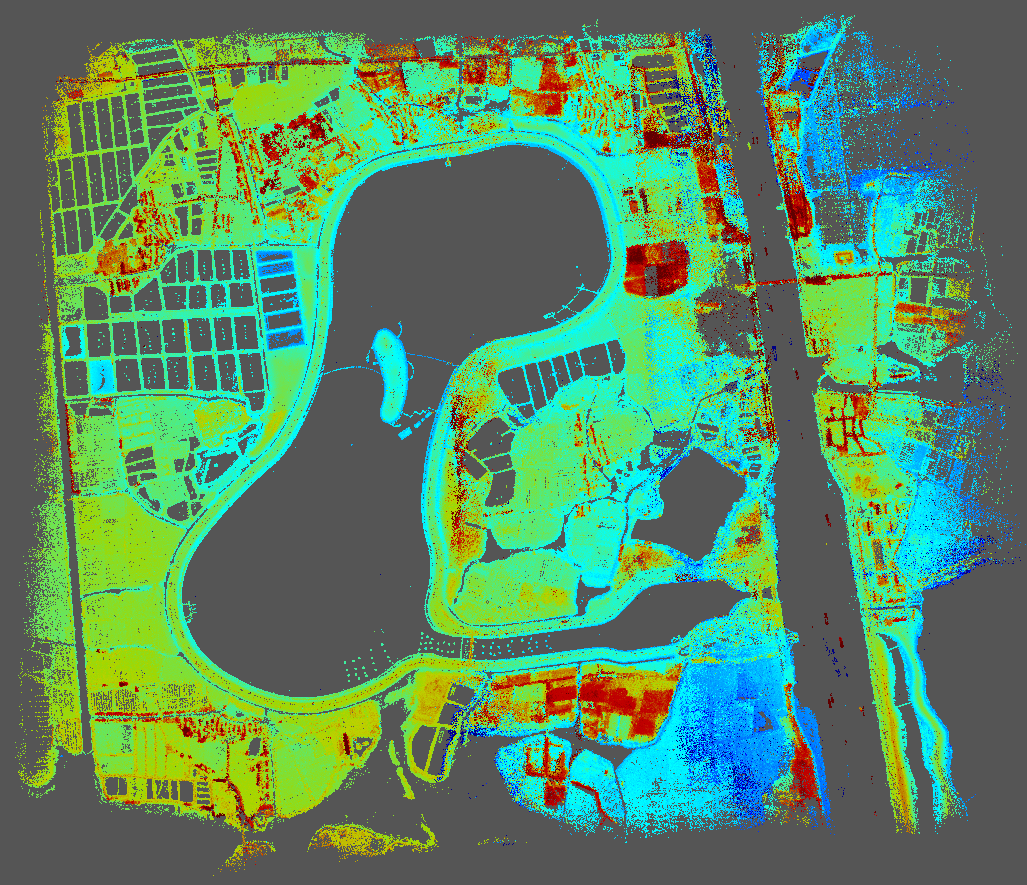}
\end{minipage}
\begin{minipage}{0.99\linewidth}
\includegraphics[width=2.7cm, height=2.5cm]{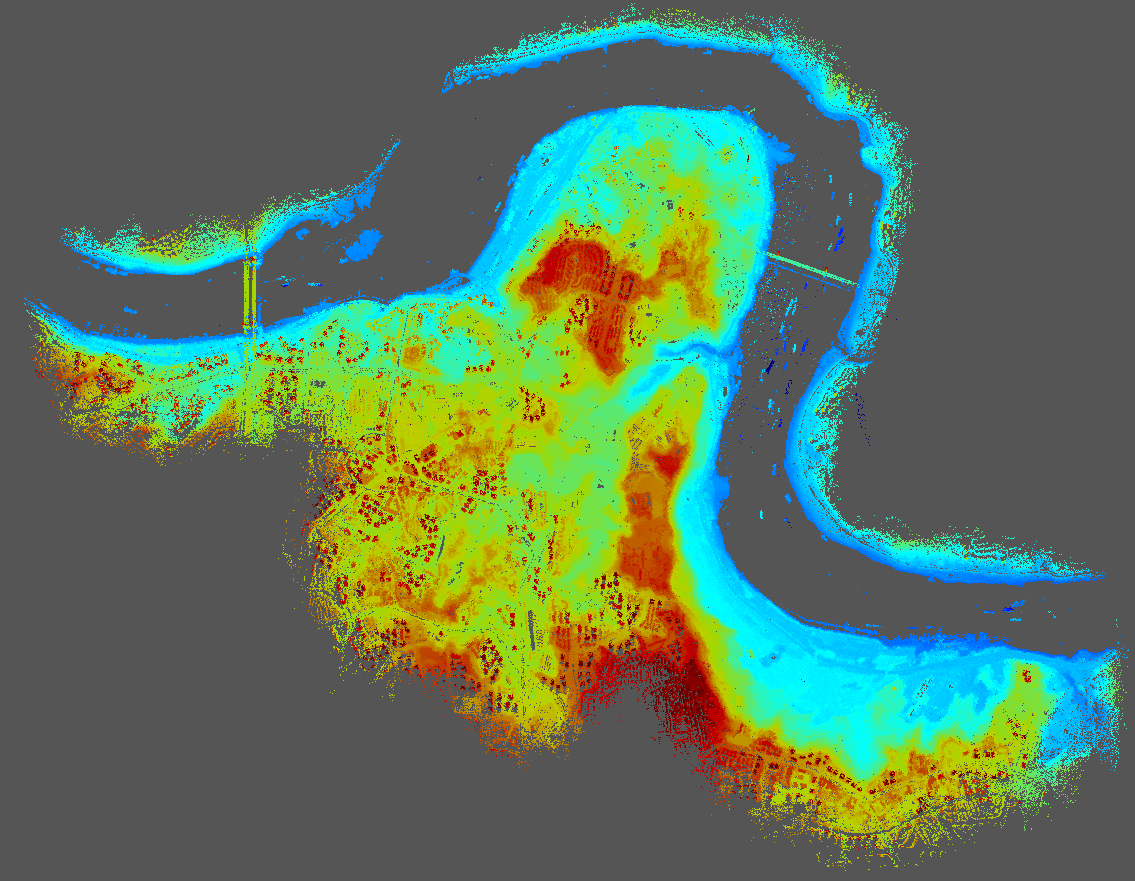}
\includegraphics[width=2.7cm, height=2.5cm]{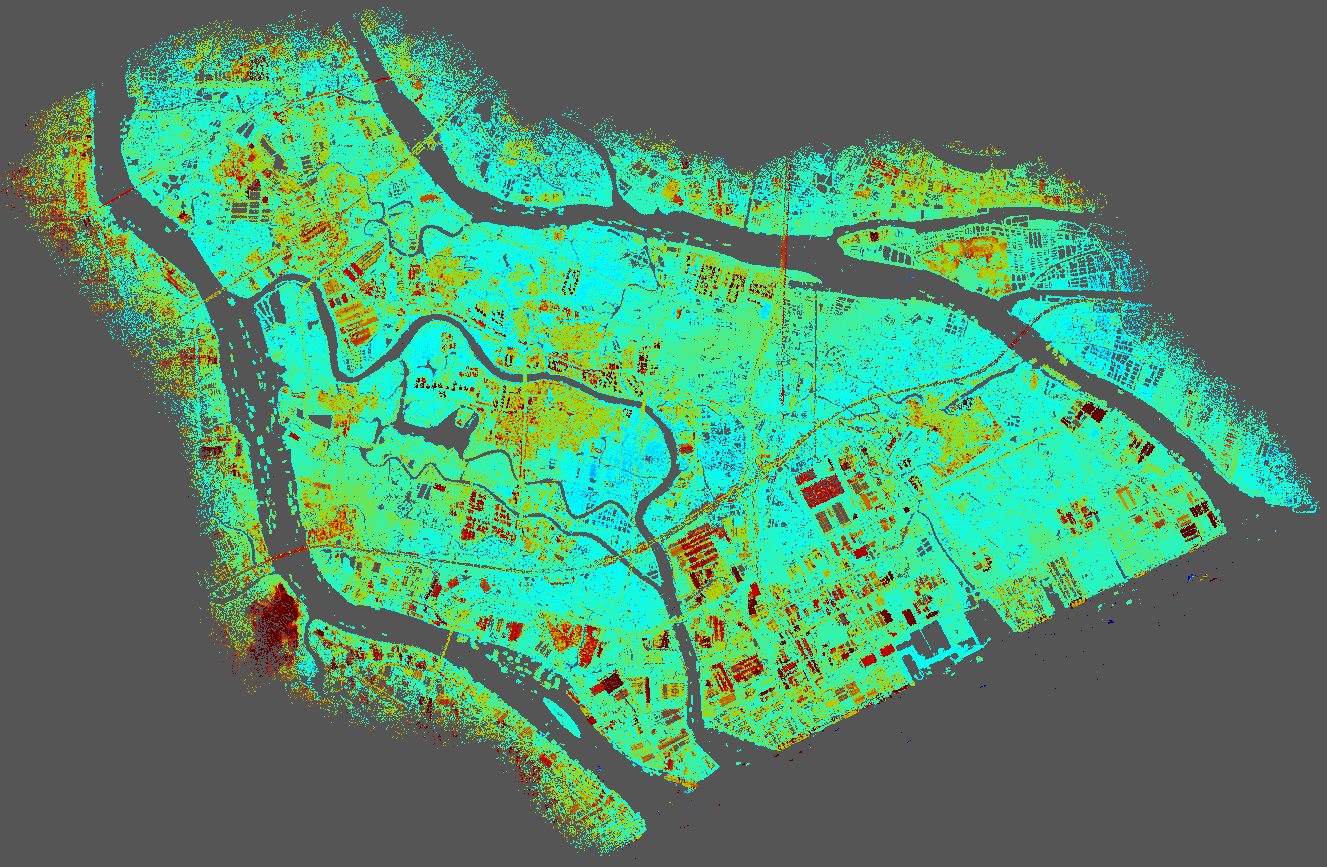}
\includegraphics[width=2.7cm, height=2.5cm]{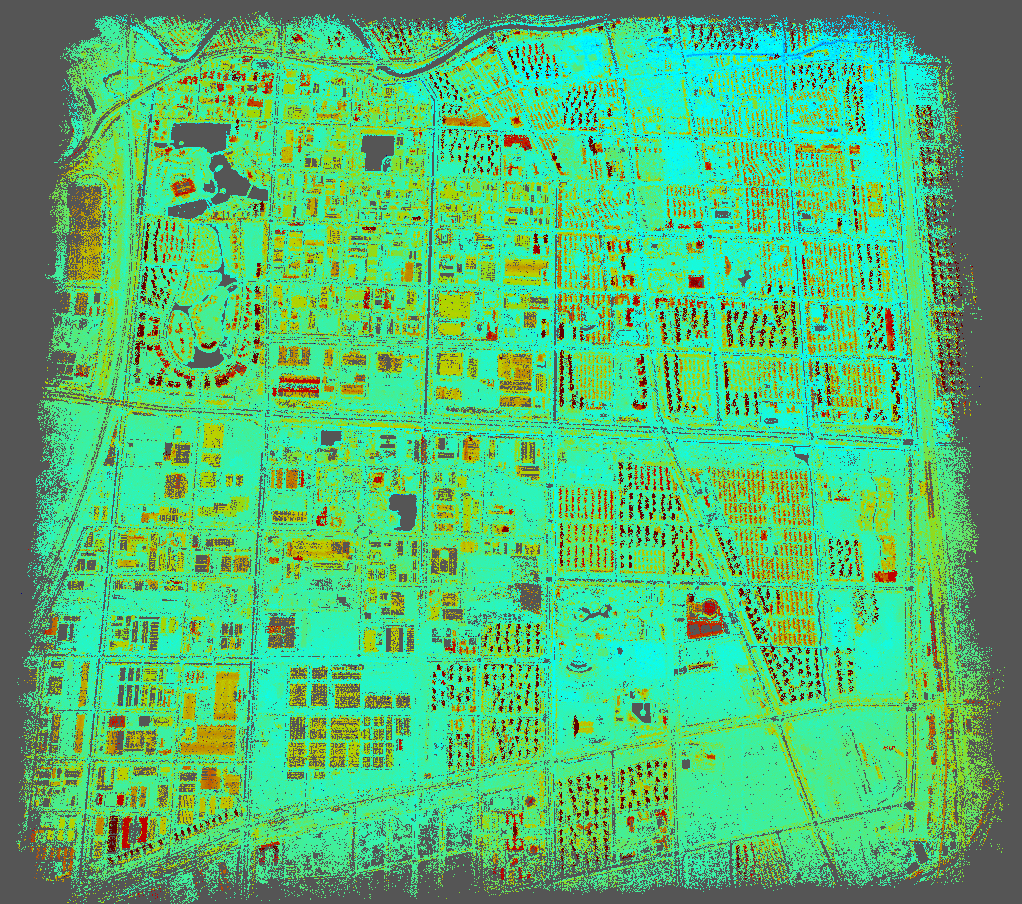}
\end{minipage}
\begin{minipage}{0.99\linewidth}
\includegraphics[width=4.67cm, height=2.5cm]{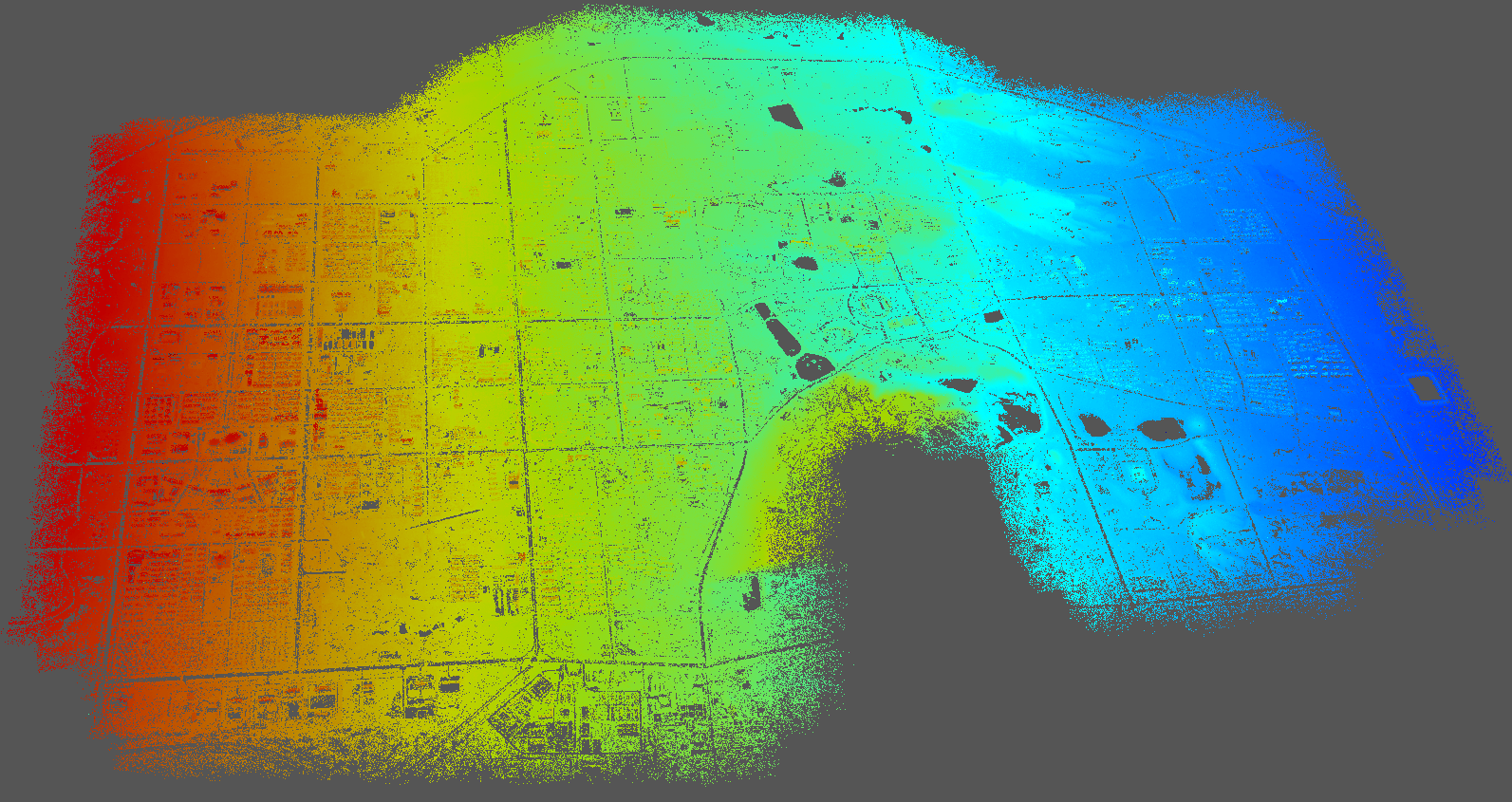}
\includegraphics[width=3.5cm, height=2.5cm]{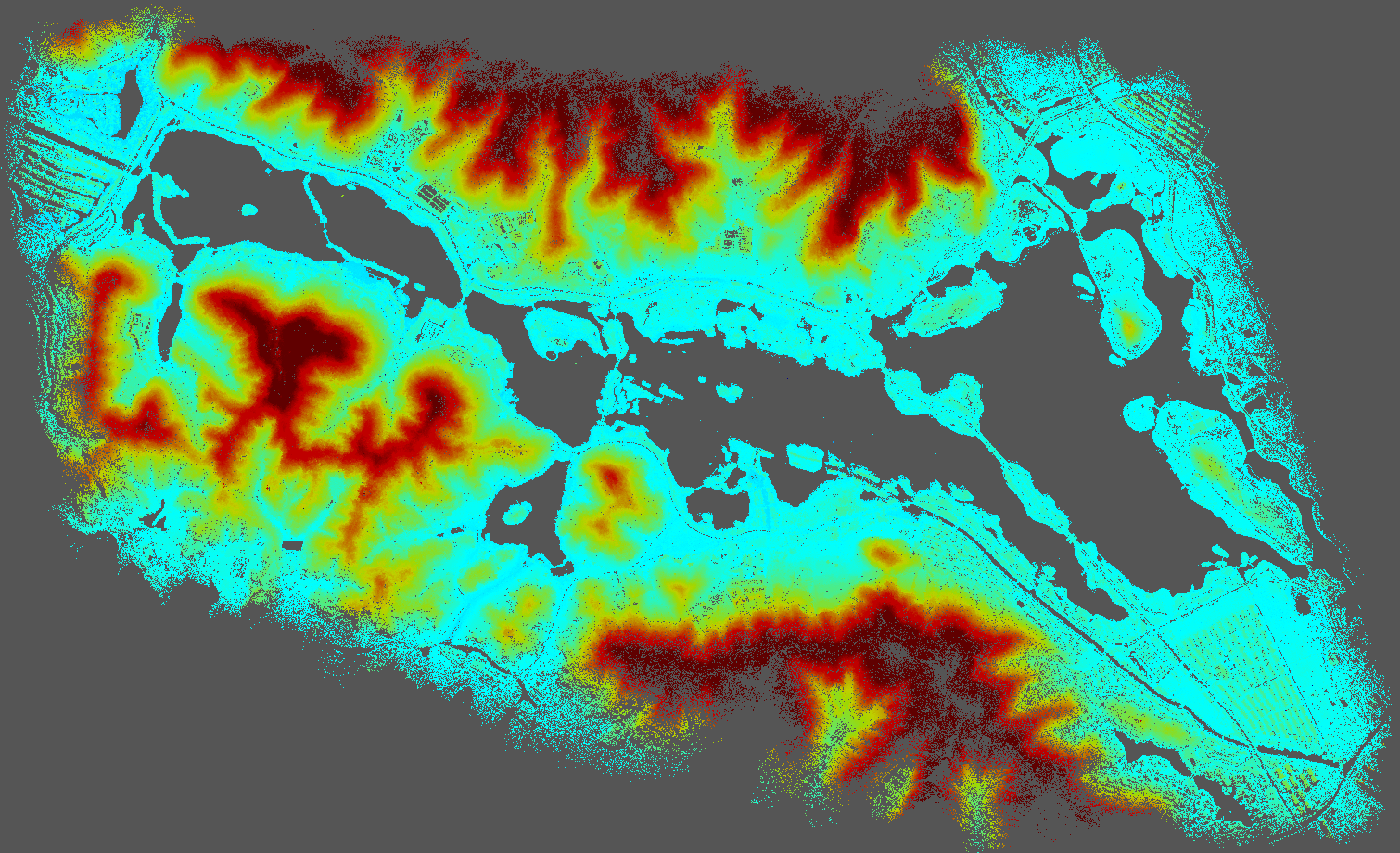}
\end{minipage}
\end{center}
   \caption{Visualization of the real datasets processed using the proposed method. From left to right, then top to bottom: SW, JLP, YQC, JZ, XS, DG, NM, MCZ, WQ, and DaPu, respectively.}
\label{fig15}
\end{figure}

\subsection{Accuracy and scalability}
\label{section7.3}
As shown in Table \ref{tab3}, The accuracy of the proposed method is almost the same as the LM algorithm applied in Ceres, which is expected because the proposed method uses the exact LM algorithm. The difference is that the tasks of forming the RCS and computing the PCG are executed in a distributed way. MegBA achieves the best accuracy, which is unexpected since it also applies the exact LM algorithm and the difference is only that it is executed on GPU. The possible reason could be the strict outlier elimination strategy used in their method. 

For the UAV datasets, the proposed method achieves about 1.2 pixels for most of datasets, and sub-pixel accuracy for two datasets. The sparse point clouds of the public datasets and the UAV datasets after bundle adjustment with the proposed method are shown in Figure \ref{fig14} and \ref{fig15}, respectively.
The proposed method successfully processed the super-large datasets (Syn1 and Syn2), and the accuracies are reasonable according to the errors added in the synthetic data, demonstrating the proposed method's vast scalability. To our own knowledge, datasets of this magnitude have never been handled by other LM-based methods. The proposed method overcomes the traditional LM algorithm's scalability bottleneck issue and relieves its limitations on image numbers by exploiting distributed computing and a memory-efficient compression format BSMC. 

\subsection{Ablation study for different parallel task assignments and node numbers}
\begin{figure}
	\begin{center}
		\includegraphics[width=0.9\linewidth]{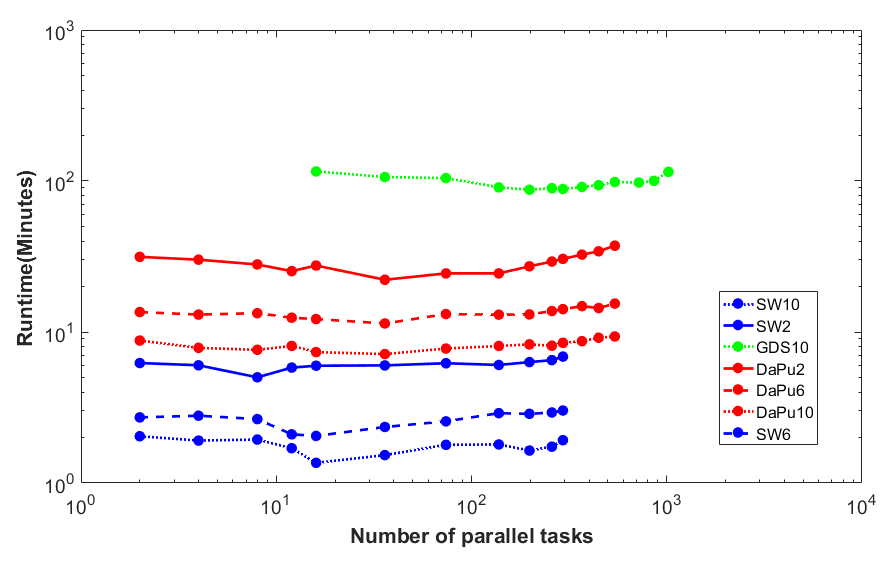}
	\end{center}
	\caption{Ablation study for different parallel task assignments and node numbers. The blue, red and green lines represent SW, DaPu and GDS datasets respectively. The solid, long dashed and short dashed lines represent the results using 2, 6 and 10 computing nodes, respectively.}
	\label{fig17}
\end{figure}

To investigate the efficiencies of different parallel task assignments and node numbers, we conduct ablation study on three datasets SW, DaPu and GDS using 2, 6 and 10 computing nodes, respectively. The GDS dataset is so large that we only test it on 10 computing nodes. The results are demonstrated in Figure \ref{fig17}. The parallel task number is neither the more the better, nor the fewer the better. For each dataset, there is a best parallel task number which achieves highest efficiency, but the best parallel task numbers various from different datasets. The best parallel task assignment scheme could be also related to the performance of the computing nodes. Nonetheless, the efficiency variation for different task assignment schemes is not big. However, it is certain that more computing nodes provide higher efficiency. The efficiency is generally linear to the number of computing nodes. 

\section{Conclusions}
\label{section8}
We parallelized the LM algorithm for bundle adjustment, which is thought to be indivisible. We found that the formation of the RCS can be divided by 3D points. So, the formation of the RCS can be executed in a distributed way using multiple computing nodes. We also introduced a block-based sparse matrix compression format to store the global RCS and sub-RCSs. The BSMC format was compared with the traditional sparse matrix compression format CSR, and our results demonstrate its superiority against the CSR on the compression rate and accessing efficiency. The other time-consuming steps in BA are also parallelized, such as the computation of PCG and the triangulation of 3D points. Based on the above, we proposed a parallel framework to conduct bundle adjustment with the exact LM algorithm in a distributed computing system. The proposed method is extensively evaluated and compared with the state-of-the-art methods using diverse datasets, including public internet datasets, real datasets collected by UAVs, and synthetic datasets. According to the experimental results and discussions, we conclude that

	(1) The proposed method is memory efficient compared with the baselines due to the application of parallel computing and the memory-efficient compression format BSMC. The accuracy is the same as the Ceres since both methods apply the LM algorithm.

	(2) The BSMC format can largely decrease the memory requirement of the RCS by 1.5 to 3.0 times compared with the traditional sparse matrix compression format CSR. The global RCS and sub-RCSs can be stored in a distributed manner based on the BSMC format.

	(3) The vast scalability of the proposed method is verified by large-scale datasets with up to 10 million images. This magnitude of dataset has never been tested by other LM-based BA methods.

The parallel mode of the proposed method is suggested for large-scale datasets. However, the improvement on the small datasets (image number less than 2K) is not obvious. The serial mode of the proposed method is suitable for the small datasets since the BSMC format works for all datasets as long as they are sparse. A limitation of the proposed method is that the global RCS, though compressed by BSMC, is still stored on the main node, which means the proposed method will eventually fail as the dataset continues to scale up. We plan to not aggregate the global RCS on the main node at all, and direct conduct the PCG in a distributed way. This will be further studied in the following research.

\section*{Acknowledgement}
This work is supported by the Special Fund of Hubei Luojia Laboratory under grant 220100034.

%------------------------------------------------------------------------

{\small
\bibliographystyle{ieee_fullname}
\bibliography{DistibutedBA}
}

\end{document}